\documentclass[11pt]{article}

\usepackage[preprint]{acl}

\usepackage{times}
\usepackage{latexsym}

\usepackage[T1]{fontenc}

\usepackage[utf8]{inputenc}

\usepackage{microtype}

\usepackage{inconsolata}

\usepackage{graphicx}

\usepackage{enumitem}
\usepackage{booktabs}
\usepackage[tikz]{bclogo}
\usepackage{amsmath}
\usepackage{amssymb}
\usepackage{multirow}
\usepackage{makecell}
\usepackage{tocloft}
\usepackage{tcolorbox}
\usepackage{diagbox}
\definecolor{msftBlack}{RGB}{0,0,0}
\newcommand{\finding}[1]{
    \begin{bclogo}[
        couleur=msftBlack!05,
        epBord=1,
        arrondi=0.1,
        logo=\bcnote,
        marge=2,  
        ombre=true,
        blur,
        couleurBord=msftBlack!10,
        tailleOndu=1.5,  
        sousTitre={\em #1}
    ]{}
    \end{bclogo}
}

\title{Retrieval, Reward, and Training Protocols: \\What Matters in Training Search Agents?}

\author{  Yibo Zhao$^{1}$ \quad
  Zichen Ding$^{1}$ \quad
  Jiayi Wu$^{1}$\\
  {\bf Zun Wang}$^{2}$\quad
  {\bf Xiang Li}$^{1}$\thanks{Corresponding Author: \texttt{xiangli@dase.ecnu.edu.cn}} \\
  $^1$School of Data Science and Engineering, East China Normal University \\
  $^2$Shanghai AI Laboratory}

\begin{document}
\maketitle
\begin{abstract}
Search agents powered by large language models can autonomously decompose queries, retrieve information, and synthesize answers through multi-step reasoning. However, the rapid growth of training methods has outpaced controlled comparison: existing works differ in retrieval corpora, reward designs, and training protocols, making it unclear what actually drives improvements. We present a controlled empirical study that isolates three under-explored dimensions of search agent training. First, we identify a critical data-coverage issue in the widely used Wikipedia 2018 corpus and show that correcting it alone yields larger gains than the differences between training algorithms. Second, we systematically compare outcome-based and process-based reward methods across three base models, finding that the simplest outcome-based approach achieves competitive or superior performance in most settings, and that process-level credit assignment can over-correct agent behavior. Third, we analyze training data diversity, off-policy data utilization, and search budget scaling, distilling practical guidelines for training effective search agents. Our code is available at \url{https://github.com/YiboZhao624/SearchAgentReview}.
\end{abstract}

\section{Introduction}

Large language models (LLMs) have advanced rapidly in recent years~\citep{flashattention, vllm, Qwen3}, demonstrating remarkable capabilities in machine translation~\citep{translation-2, translation-1, translation-3}, reasoning~\citep{reasoning-3, reasoning-1, reasoning-2}, and creative writing~\citep{writing-3, writing-2, writing-1}. More recently, LLMs have evolved beyond standalone models into autonomous agents~\citep{chatbot2agent} capable of interacting with external environments, giving rise to computer-using agents~\citep{cua2025, scalecua, yang2026symphony}, coding agents~\citep{coding-agent, kimi, coding-agent-2}, and search agents~\citep{deep-search-2, deep-search-1}.

As LLM-based agents grow increasingly capable, rigorous evaluation and comparison of different training approaches becomes essential to guide future research. For computer-using and coding agents, execution is grounded in a shared sandbox (e.g., Docker containers) that normalizes the action-execution interface, leaving limited room for variation and enabling convergence on standardized evaluation protocols~\citep{swebench, osworld}. As a result, the community has largely converged on standardized evaluation protocols~\citep{swebench, osworld}. Search agents, however, face a far less constrained design space: the retrieval source, tool interface, action space, and granularity of retrieval vary freely, with no community standard in sight. 
Although comprehensive benchmarks such as GAIA~\citep{GAIA} and BrowseComp~\citep{browsecomp} exist, they standardize only the evaluation questions without providing a shared training environment for comparing training methods. 
In practice, researchers always fall back on conventional multi-hop QA benchmarks~\citep{2wiki, musique, hotpotqa} with local retrieval,  where differences in tool design, reward formulation, and hyperparameter choices across studies make reliable cross-method comparison difficult.

As a result, despite the rapid growth of research on search agents, the community has not yet reached consensus on fundamental questions such as what drives improvements.
In this work, rather than proposing a new algorithm, we aim to provide such answers through a controlled empirical study. 
Specifically, while recent works have extensively studied algorithmic aspects of reinforcement learning (RL) training for search agents, such as dynamic filtering, importance sampling clipping, and entropy control for stabilizing policy optimization~\citep{llds, arlarena,ragen2}, these analyses primarily address \textit{how to optimize reliably}. In contrast, equally fundamental questions about \textit{what the agent learns from}, including reward design, credit assignment, and training data composition, remain largely underexplored, and existing works have made divergent decisions along these dimensions without understanding their effects.

We present a controlled empirical study of RL training for search agents. Our contributions are:

\begin{itemize}[leftmargin=*]

\item \textbf{Retrieval Environment.} We identify a critical yet previously overlooked issue in the widely adopted Wikipedia 2018 corpus~\citep{wiki18}: a significant portion of relevant passages are missing, causing retrieval failures and spurious training signals. We construct a more complete retrieval corpus and show that this correction alone yields larger performance gains than the differences between training algorithms, underscoring that retrieval environment quality is a prerequisite for reliable comparison.


\item \textbf{Reward Design and Credit Assignment.} We systematically benchmark three 
process reward methods and one outcome reward method across three base models. 
Our results reveal that the simplest outcome-based approach achieves competitive or superior performance in most settings, 
questioning whether complex process reward designs consistently justify their added complexity.
Further analysis of intermediate search behavior shows that process-level credit assignment can over-correct agent strategies, improving one aspect of search quality at the cost of another.

\item \textbf{Data, Off-Policy Degree, and Search Budget.} We conduct a detailed analysis of training data diversity, the degree of off-policy data usage, and search budget scaling during both training and inference, distilling practical guidelines for optimizing search agent performance.

\end{itemize}

Together, these contributions provide the community with controlled empirical insights and practical guidelines for training search agents under a unified and fair experimental setup.



\section{Related Work}

\subsection{Reward Design for Search Agent}\label{sec:reward_design}

Existing reward designs for search agents range from trajectory-level outcome rewards to step-level process rewards that aim for finer-grained credit assignment. Search-R1~\citep{searchr1}, and R1-Searcher~\citep{r1searcher} represent the outcome-reward paradigm: a rule-based verifier scores the model's final answer against the ground truth using metrics such as exact match or token-level F1. However, such trajectory-level signals provide no supervision for individual retrieval steps. 

To alleviate the sparsity of trajectory-level rewards, recent work estimates step-level credit through three broad paradigms, distinguished by how they construct the training signal.

\textbf{Structural comparison} methods build explicit branching structures and derive preference pairs from sibling nodes. ReasonRAG~\citep{reasonrag} rolls out multiple trajectories from a stronger model to construct off-policy DPO preference data with an analogous tree-like structure.
Tree-GRPO~\citep{tree-grpo} expands on-policy search trajectories step by step into a tree structure, using sibling outcomes as natural contrastive pairs while reducing the budgets of tool calls.

\textbf{Cross-trajectory aggregation} compares steps across independent rollouts without explicit structure. 
GiGPO~\citep{gigpo} groups steps from different trajectories by matching intermediate states via text similarity, then computes subgroup stepwise advantages from this post-hoc grouping.

\textbf{Information-theoretic} methods model search as progressively gathering information toward the ground truth, and score each step by a proxy of its information gain.
StepSearch~\citep{stepsearch} uses a stronger model to generate sub-queries, then uses their retrieval results as a reference signal to approximate the information gain of each step.
IGPO~\citep{igpo} measures the change in the model's likelihood of producing the correct answer before and after a retrieval step.

However, each method reports results under its own corpus and configuration. Without a controlled setup that isolates reward design from these variables, it is unclear whether gains reflect better credit assignment or favorable evaluation conditions.

\subsection{Understanding Training Instability in RL for Search Agents}

A complementary line of research has focused on diagnosing \textit{why} RL training of search agents is prone to instability. 
LLDS~\citep{llds} identifies that high overlap between positive and negative trajectories in tool-use actions causes gradient updates to inadvertently suppress correct behaviors, leading to training collapse. 
RAGEN-2~\citep{ragen2} attributes collapse to model outputs degenerating into a fixed, question-agnostic template that overwhelms the gradient signal with noise.
ARL-Arena~\citep{arlarena} studies the effects of importance sampling, loss aggregation, and advantage computation on training stability. 
Calibadv~\citep{calibadv} attributes training instability to imbalanced positive and negative advantages under coarse-grained credit assignment.

While these works focus on why training fails at the optimization level, the effects of retrieval environment, reward design, and training data composition have not been compared under a unified setup that controls for confounding factors. Our work provides this controlled comparison.

\section{Experiments Setup}

\subsection{Training Algorithms}

Following the taxonomy in Sec.~\ref {sec:reward_design}, we select four methods spanning four credit assignment strategies: Search-R1~\citep{searchr1}, the most widely adopted outcome-reward baseline without heuristic credit assignment; GiGPO~\citep{gigpo}, for cross-trajectory aggregation; and Tree-GRPO~\citep{tree-grpo} and IGPO~\citep{igpo}, for structural comparison and information-theoretic scoring, respectively, both free of dependence on strong models.
All four are re-implemented within a shared GRPO~\cite{reasoning-1} objective to isolate the effect of credit assignment.

Given a query $q$, GRPO samples a group of $G$ trajectories $\{\tau^{(1)}, \ldots, \tau^{(G)}\}$ from the current policy $\pi_\theta$. Each trajectory receives a final reward $r^{(i)}$. 
The advantage $\hat{A}^{(i)}$ of each trajectory is computed by normalizing rewards within the group. 
The policy is then updated by maximizing the asymmetric clipped surrogate objective~\citep{dapo}:
\begin{align}
\mathcal{J}(\theta)
&= \mathbb{E}\Bigg[
\frac{1}{\sum_{i=1}^{G} |\tau^{(i)}|}
\sum_{i=1}^{G} \sum_{t=1}^{|\tau^{(i)}|}
\min\Big(
\rho_t^{(i)} \hat{A}^{(i)}, \nonumber\\
&
\operatorname{clip}\big(\rho_t^{(i)}, 1-\epsilon_{\text{low}}, 1+\epsilon_{\text{high}}\big)\hat{A}^{(i)}
\Big)
\Bigg],
\end{align}
where $\rho_t^{(i)}$ is the importance ratio, defined as ${\pi_{\theta}(a_t^{(i)}\mid h_t^{(i)})}/{\pi_{\theta_\text{old}}(a_t^{(i)}\mid h_t^{(i)})}$.
Due to page limitations, we defer full algorithmic details to App.~\ref{app:algorithm} and briefly describe each method's core heuristic here:
\begin{itemize}[leftmargin=*]

\item \textbf{Search-R1} uses outcome reward (EM) with no step-level credit: all tokens share a single trajectory-level advantage.

\item \textbf{GiGPO} groups steps across trajectories by state similarity, enabling step-level advantage by comparing actions from matched states.

\item \textbf{IGPO} uses per-turn change in log-probability of the ground truth as a heuristic step-level reward.

\item \textbf{Tree-GRPO} expands intermediate nodes to produce prefix-sharing trajectory pairs, creating natural step-level comparisons at branch points.

\end{itemize}

\subsection{Experiment Settings}

To ensure a fair comparison, we randomly sample a combined total of 9,000 training instances from HotpotQA~\cite{hotpotqa}, MuSiQue~\cite{musique}, and 2WikiMultihopQA~\cite{2wiki} as the unified training set. 
All approaches are implemented in the same Verl~\cite{verl} codebase to control for infrastructure differences. 
For evaluation, we sample up to 1,000 test instances from each of HotpotQA, 2WikiMultihopQA, MuSiQue, Bamboogle~\cite{bamboogle}, and PopQA~\cite{popqa}, resulting in 4,125 test instances in total. 
Unless otherwise specified, the base model is Qwen3-8B~\cite{Qwen3} with the Hermes-format tool-calling interface~\cite{hermes}. 
Other defaults are: a maximum of 4 tool-call turns, a batch size of 32, a mini-batch size of 16, a maximum response length of 4096, and one training epoch. 
Subsequent experiments vary one factor at a time from this default.
We use a decoding temperature of 0.6 and report the mean@4 Exact Match (EM) performance to reduce evaluation variance. 
A complete list of hyperparameters is in App.~\ref{app:hyper}.

\section{Analysis}
We investigate five factors that affect search agent training, organized into three groups: the retrieval corpus (Sec.~\ref{sec:corpus}), the reward design (Sec.~\ref{sec:reward}), and the training protocol (Sec.~\ref{sec:protocol}). Due to space limitations, most experimental results are presented as figures. Detailed numerical results and per-dataset results are provided in App.~\ref{app:detailed_res}, and all the trained models are released at \url{https://hf.co/collections/ybyby624/search-agent-review}.

\subsection{Retrieval Corpus}\label{sec:corpus}

\begin{figure}[t]
    \centering
    \includegraphics[width=\linewidth]{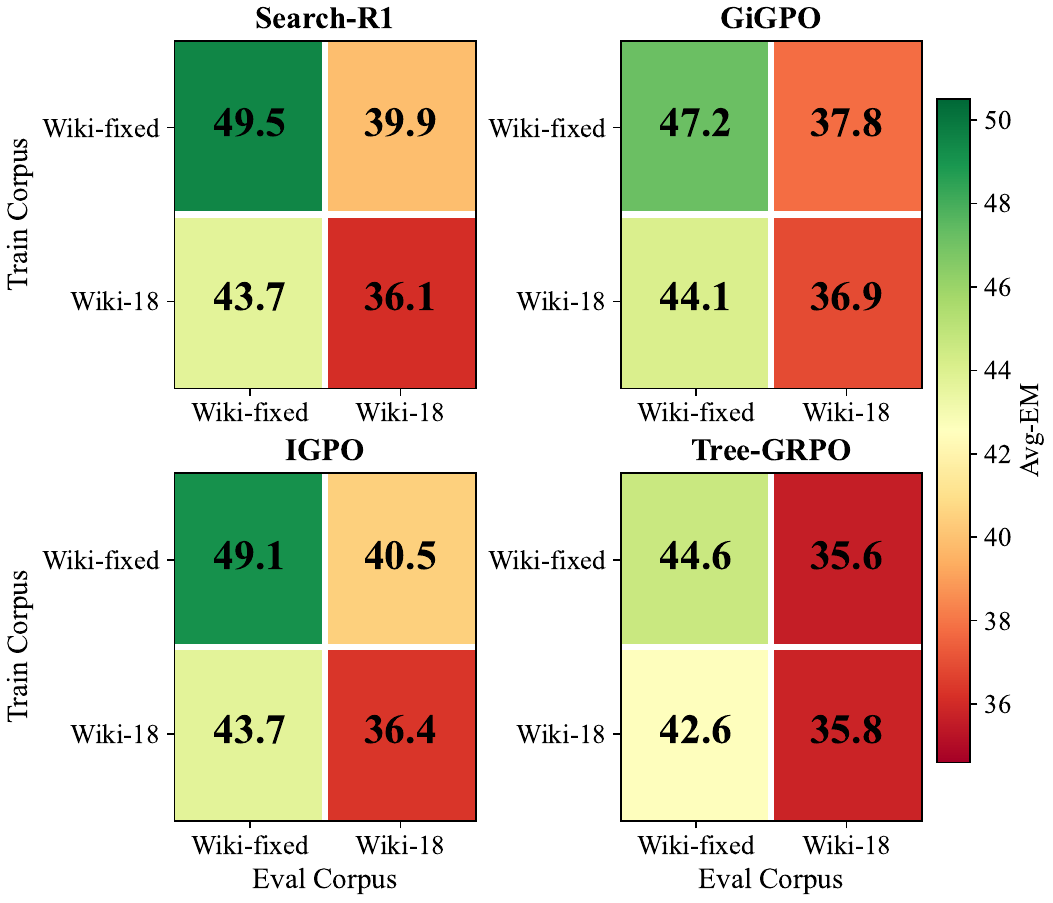}
    \caption{Effect of search environment on average EM across five benchmarks. Each cell shows the Avg. EM for a given train/test environment combination.}\label{fig:env}
\end{figure}

The widely used Wiki-18 corpus~\cite{wiki18} lacks many documents required for multi-hop reasoning. 
We compared all annotated supporting documents from the training and validation splits of HotpotQA, 2WikiMultihopQA, and MuSiQue against Wiki-18 and found that \textbf{295,331 supporting documents are absent}.
Among our 9,000 training instances, 3,321 correspond to questions whose gold evidence simply cannot be retrieved, making them inherently unanswerable under this retrieval corpus.

At first glance, these missing-document questions might seem harmless: if the model cannot retrieve the gold documents, all rollouts within a group would receive zero reward, producing no gradient signal. 
However, inspection of the training rollouts tells a different story. 
For Search-R1, the model answered 1,697 of the 3,321 unanswerable questions correctly at least once, and for 831, every rollout in the group produced the correct answer. This means 866 training instances, roughly \textbf{10\% of the full training set}, 
generate gradient signals that may largely reflect correct guesses from parametric memory rather than successful retrieval, potentially introducing noise into the optimization of the search policy.
Similarly, GiGPO, IGPO, and Tree-GRPO exhibit 890, 698, and 859 such noise-prone groups, respectively, confirming that this issue is systematic across different methods and not an artifact of any particular training algorithm.

To remove this confound, we augment Wiki-18 by adding all missing supporting documents, constructing a revised corpus we call \textbf{Wiki-fixed}.

We retrain all four methods on both the original and revised corpora. 
As shown in Fig.~\ref{fig:env}, training on Wiki-18 yields 2--6 EM points lower performance than training on Wiki-fixed under identical evaluation. 
\textbf{This gap even exceeds the performance differences among different training algorithms}, indicating that corpus completeness is a more decisive factor than reward design in this setting.
The incomplete corpus also obscures algorithmic comparison in two aspects: 
(1) methods that are clearly separable under Wiki-fixed converge to near-identical performance when trained and evaluated on Wiki-18, and 
(2) method rankings shift, Search-R1 drops from first place under Wiki-fixed to third under Wiki-18, suggesting that evaluations under an incomplete corpus may obscure the true performance differences between methods.

\begin{table}[tbp]
\centering
\resizebox{0.85\linewidth}{!}{\begin{tabular}{cccc}
\toprule
\diagbox[]{Method}{Train Env.} & Wiki-18 & Wiki-fixed & $\Delta$\\
\midrule
Search-R1 & 45.81 & 46.09 & 0.28\\
GiGPO & 46.03 & 45.81 & 0.22 \\
IGPO & 46.84 & 47.87 & 1.03\\
Tree-GRPO & 44.89 & 44.00 & 0.89\\
\bottomrule
\end{tabular}}
\caption{The average EM across five benchmarks tested with the Wiki-fixed corpus. $\Delta$ reports the absolute difference between Wiki-18 and Wiki-fixed when training on entries answerable under both corpora.}
\label{tab:compare_corpus}
\end{table}

To further isolate the cause, we exclude all missing-document questions from the training data and retrain on both corpora. 
As shown in Tab.~\ref{tab:compare_corpus}, when every training instance is answerable, the gap between Wiki-18 and Wiki-fixed becomes negligible, confirming that the missing documents, not other differences, are the root cause. 

\finding{The retrieval environment matters more than the training algorithm choice. Ensuring that all training questions are answerable under the given corpus is a prerequisite for reliable comparison of training methods.}

\subsection{Reward Design}\label{sec:reward}

We evaluate reward design from two angles: outcome evaluation, which measures final answer accuracy, and process evaluation, which examines the quality of intermediate search behavior.

\subsubsection{Outcome Evaluation}

\begin{table*}[htbp]
\centering
\resizebox{0.95\textwidth}{!}{\begin{tabular}{cccccccccc}
\toprule
Model& Format & Method & 2Wiki & Bamboogle & HotpotQA & Musique & PopQA & Avg. EM & Avg. Turns \\
\midrule
\multirow{4}{*}{Qwen3-8B} & \multirow{4}{*}{Hermes} 
   & Search-R1 & \textbf{70.55} & 47.00 & \textbf{57.92} & \textbf{27.02} & 42.80 & \textbf{49.49} & 1.72\\
 & & GiGPO     & 67.95 & \textbf{52.00} & 53.37 & 24.37 & 42.65 & 47.23 & 1.62\\
 & & IGPO      & 70.32 & 47.40 & 56.72 & 25.05 & \textbf{44.62} & 49.12 & 2.99\\
 & & Tree-GRPO & 65.27 & 43.80 & 51.15 & 21.87 & 40.32 & 44.63 & 2.07 \\
\midrule
\multirow{4}{*}{\makecell{Qwen2.5-\\7B-Instruct}} & \multirow{4}{*}{Hermes} 
   & Search-R1 & 66.02 & \textbf{45.80} & \textbf{52.80} & \textbf{21.55} & 40.35 & \textbf{45.20} & 2.09 \\
 & & GiGPO     & \textbf{67.60} & 41.60 & 50.00 & 20.32 & \textbf{40.80} & 44.58 & 1.58 \\
 & & IGPO      & 44.80 & 33.80 & 44.37 &  9.85 & 36.07 & 33.77 & 2.72\\
 & & Tree-GRPO & 57.17 & 38.60 & 48.85 & 19.80 & 41.85 & 41.81 & 2.01 \\
\midrule
\multirow{4}{*}{Qwen3-8B} & \multirow{4}{*}{Search-R1} 
   & Search-R1 & 59.90 & 49.40 & 51.45 & 19.90 & 37.50 & 42.40 & 2.13\\
 & & GiGPO     & 58.60 & \textbf{52.45} & 51.74 & \textbf{25.00} & \textbf{45.56} & \textbf{45.44} & 1.61 \\
 & & IGPO      & 58.87 & 48.40 & 52.05 & 19.80 & 40.10 & 42.87 & 1.92\\
 & & Tree-GRPO & \textbf{63.65} & 47.80 & \textbf{52.07} & 22.80 & 40.62 & 44.87 & 2.08 \\
\midrule
\multirow{4}{*}{\makecell{Qwen2.5-\\7B-Instruct}} & \multirow{4}{*}{Search-R1} 
   & Search-R1 & 57.32 & \textbf{40.80} & 47.67 & 18.22 & 35.02 & 39.60 & 2.11 \\
 & & GiGPO     & 53.17 & 38.80 & 48.17 & \textbf{21.50} & \textbf{42.30} & 41.21 & 1.67 \\
 & & IGPO      & 47.35 & 39.60 & 48.25 & 18.32 & 38.45 & 38.13 & 2.01\\
 & & Tree-GRPO & \textbf{59.75} & 39.00 & \textbf{48.60} & 17.70 & 39.30 & \textbf{41.26} & 1.88 \\
\midrule
\multirow{4}{*}{\makecell{Qwen2.5-\\7B-Base}} & \multirow{4}{*}{Search-R1} 
   & Search-R1 & \textbf{63.42} & 42.60 & \textbf{52.57} & \textbf{25.20} & 41.25 & \textbf{45.52} & 1.99 \\
 & & GiGPO     & 58.95 & \textbf{44.00} & 49.22 & 22.82 & 43.12 & 43.54 & 1.61 \\
 & & IGPO      & 34.95 & 23.20 & 34.50 &  6.90 & \textbf{43.27} & 29.70 & 1.00 \\
 & & Tree-GRPO & 57.50 & 40.40 & 46.42 & 17.52 & 40.72 & 40.53 & 1.74 \\
\bottomrule
\end{tabular}}
\caption{Performance comparison across five datasets, three base models, and two tool call formats. The \textbf{best results} within a comparing group are bold. Avg. Turns means the average search turns called by the agent.}
\label{tab:method_comparison}
\end{table*}

We test the four algorithms across three different models, including Qwen3-8B\footnote{https://huggingface.co/Qwen/Qwen3-8B}, Qwen2.5-7B-Instruct\footnote{https://huggingface.co/Qwen/Qwen2.5-7B-Instruct}, and Qwen2.5-7B-Base\footnote{https://huggingface.co/Qwen/Qwen2.5-7B};
and two different tool-call formats, including Hermes and Search-R1 format (XML style). 
Detailed prompt templates are provided in App.~\ref{app:prompt}. 

As illustrated in Tab.~\ref{tab:method_comparison}, under controlled comparison, the simplest outcome-based method achieves the highest average EM in three of five settings, demonstrating that simple outcome-based reward remains a strong baseline. 
GiGPO performs consistently well across settings and achieves the best result in one setting (Qwen3-8B with Search-R1 format), suggesting that sub-group advantage is a reliable approach to process-level credit assignment.
Tree-GRPO achieves above 40 average EM in all five settings, exhibiting the most stable performance among the four methods. However, its ceiling remains low: it never exceeds 45 points, and it achieves the best average only in one setting (Qwen2.5-7B-Instruct with Search-R1 format), where the other methods also cluster near this range. This stability-without-peak pattern suggests that tree-expansion comparison provides a conservative learning signal that avoids catastrophic failures but lacks the capacity to push performance further.
In contrast, IGPO shows the largest gap between its upper and lower performance bounds across settings. 
Under Qwen3-8B with Hermes format, IGPO achieves an average EM of 49.12, whereas on Qwen2.5-7B-Base it reaches only 29.70. 
We attribute this to the fact that, on the base model, the model’s in-context learning and instruction-following abilities are comparatively weak, making the heuristic based on the log probability of generating the correct answer ineffective. Instead of providing useful guidance, it may introduce substantial noise, preventing the model from learning meaningful signals.
By contrast, on the stronger Qwen3-8B model, the log probability can more accurately reflect the incremental information brought by retrieval, leading to much better performance. 

\finding{Heuristic process credit assignment does not consistently outperform outcome-level supervision under controlled comparison.}

From the experimental results, we further derive that the tool call format is highly important. Training Qwen3-8B or Qwen2.5-7B-Instruct with the Hermes format, consistent with their post-training setup, leads to better performance than using an XML-based format, such as the Search-R1 style defined through a system prompt.
At the same time, applying reinforcement learning directly to the base model  yields higher performance than instruction-tuned counterparts (Qwen2.5-7B-Instruct and even Qwen3-8B), provided the tool-call format is kept simple enough for the base model to learn from scratch.
We also experimented with three different methods for applying the Hermes-format tool call to Qwen2.5-7B-Base. However, because the base model was unable to reliably generate JSON-formatted text, the training completely failed.

\finding{Future work should keep the tool call format consistent when comparing against baselines to attribute gains to the algorithm.}


\subsubsection{Process Evaluation}\label{sec:process}

The findings above reveal how credit assignment affects final performance, but do not explain the intermediate search behavior.
As shown in the Avg. Turns column of Tab.~\ref{tab:method_comparison}, search depth varies substantially across methods: under the Hermes format on Qwen3-8B, IGPO averages nearly twice as many turns as GiGPO. 
This gap motivates a closer look at intermediate search behavior: whether different credit assignment strategies shape distinct search patterns, and whether training actually improves search capability over the untrained counterpart.


\begin{table}[t]
\centering
\resizebox{\linewidth}{!}{\begin{tabular}{lcccc}
\toprule
Method & Search-R1 & GiGPO & IGPO & Tree-GRPO \\
\midrule
Avg. EM & 63.98 & 63.50 & 66.08 & 59.50 \\
Avg. Turns & 2.00 & 1.67 & 3.00 & 2.05 \\
\midrule
R {\small trained}\quad$\uparrow$ & 37.23 & \textbf{49.58} & 43.09 & 39.76 \\
R {\small untrained}\quad$\uparrow$ & \textbf{41.73} & 40.00 & \textbf{43.62} & \textbf{42.55} \\
\midrule
O {\small trained}\quad$\downarrow$ & \textbf{4.20} & 10.48 & 40.15 & 26.51 \\
O {\small untrained}\quad$\downarrow$ & 7.87 & \textbf{7.64} & \textbf{36.02} & \textbf{8.66} \\
\bottomrule
\end{tabular}}
\caption{Process evaluation results. R stands for recall rate, and O stands for overlap rate. The better result between the trained and untrained counterparts is bold.}\label{tab:process}
\end{table}

To answer these questions, we sample 2,000 unused examples from our training set, each with annotated supporting documents, and roll out the three Hermes-format Qwen3-8B models. 
After we obtain the trajectories, we decompose them into individual search steps. 
At each step, we feed the accumulated search history to the untrained model and explicitly prompt it to generate the next search query based on the question and the information collected so far. 
We then compare the two sets of retrieved documents on two metrics: recall of supporting documents and overlap with documents retrieved in earlier steps.

The experimental results are summarized in Tab.~\ref{tab:process}. We find that models trained with four algorithms exhibit clearly different preferences. In terms of the number of search turns, IGPO tends to encourage more extensive searching, GiGPO tends to encourage fewer searches, and Search-R1 and Tree-GRPO lie in between.
In terms of query recall, comparing the trained models with their untrained counterparts reveals that 3 of 4 methods \emph{decrease} recall, suggesting that search agents are not necessarily better query writers.
Search-R1 and Tree-GRPO drop by 4.50 and 2.79 points, respectively, while IGPO shows only a slight decrease ($-0.53$). 
In contrast, GiGPO improves by nearly 10 points.
The overlap rate offers another perspective: Search-R1 learns to retrieve documents different from previously seen ones, while the other three methods increase redundancy.

Beyond the trained-vs-untrained comparison, a consistent trade-off emerges across the four methods: models that cannot produce high-quality or diverse queries compensate by searching more turns, while those with stronger per-step retrieval are less inclined to continue. GiGPO sits at one extreme with high recall but fewest turns; IGPO sits on the other side with extensive but redundant searching. Search-R1 and Tree-GRPO fall in between.

Taken together, these results provide a more complete picture of the three methods. Search-R1 trains the model to explore in more diverse directions: although the recall at each step is not high, the overlap rate is relatively low. GiGPO improves the model's ability to generate high-quality queries, but at the same time suppresses further exploration, as reflected in its higher recall rate but fewer search turns. By contrast, IGPO encourages very thorough searching, even if part of it may be redundant. Tree-GRPO exhibits intermediate behavior across most metrics, without strongly favoring either query quality or search depth.

\begin{figure*}[t]
    \centering
    \includegraphics[width=\textwidth]{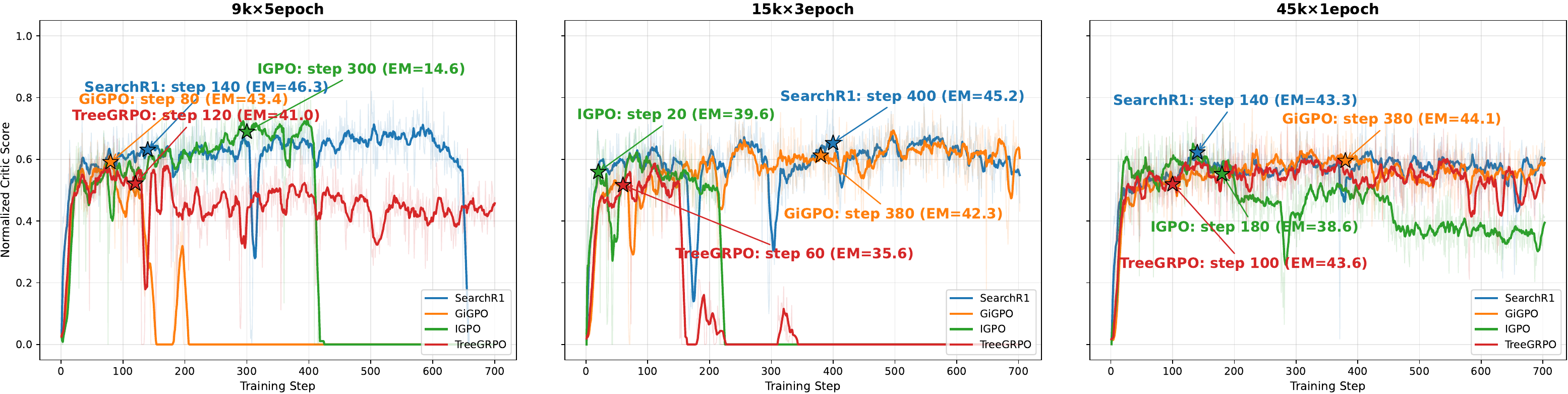}
    \caption{Training curves of four methods across three data settings. Stars indicate the best checkpoint selected based on the average exact match on the validation set.}\label{fig:curves}
\end{figure*}

We attribute these phenomena to the different credit assignment strategies. 
GiGPO computes the reward for each step using discounted return, such that actions closer to the end of the trajectory receive higher discounted rewards. This encourages the model to reduce the number of search steps, making each action closer to the trajectory end. 
In contrast, IGPO assigns an information gain reward to every step. When the final answer is incorrect or the outcome reward is close to zero, the information gain reward dominates the update direction of the entire trajectory, pushing the model toward over-searching behavior.
Tree-GRPO compares steps across different rollout branches rather than applying a per-step scalar reward. This relative signal neither penalizes additional turns as strongly as discounted return nor incentivizes every step individually, resulting in intermediate behavior without a dominant bias toward either direction.
Search-R1, by contrast, introduces no such heuristics and relies solely on the final outcome to determine the optimization direction, leaving the search strategy to emerge freely from the training signal.

\finding{Applying credit assignment to the process does not necessarily improve the intermediate trajectory itself. The stronger the heuristic signal, the more it biases the model toward a specific search pattern — sometimes at the cost of overall quality.}

\subsection{Training Protocol}\label{sec:protocol}

Having examined the retrieval corpus and reward design, we turn to three training protocol choices: data diversity (Sec.~\ref{sec:data}), off-policy degree (Sec.~\ref{sec:offpolicy}), and search budget, i.e., the maximum number of tool-call turns allowed during training and evaluation (Sec.~\ref{sec:maxturns}).

\subsubsection{Data}\label{sec:data}

Previous work varies widely in the data regime: some methods train for a single epoch on large datasets~\cite{searchr1}, while others repeat smaller datasets over multiple epochs~\cite{deep-search-2}.
The key difference is data diversity — whether the model sees more unique examples or revisits fewer ones. 
To isolate this factor, we train Qwen2.5-7B-Instruct with the Search-R1 format and fix total training compute at approximately 700 steps (batch size 64) and vary the diversity: 9,000 examples for 5 epochs, 15,000 for 3 epochs, and 45,000 for 1 epoch. 
Training curves and selected checkpoints are shown in Fig.~\ref{fig:curves}.

 Across all three data scales, the best-selected checkpoints achieve comparable final performance, indicating that greater diversity alone does not improve outcomes. 
 However, training dynamics differ markedly: with 9,000 or 15,000 instances, at least one method collapses in the latter half of training, whereas the 45,000-instance setting yields smoother curves for GiGPO and Search-R1 without spurious spikes.
 This suggests that data diversity primarily benefits training stability rather than final performance.
 Examining checkpoint selection more closely, the best checkpoints for both Search-R1 and GiGPO often emerge well before training concludes, yet achieve comparable final EM across settings.
 In most cases, performance peaks within the first half of training and degrades thereafter, indicating that prolonged training does not yield further gains and that proper checkpoint selection is essential regardless of data scale.

\subsubsection{Off-policy}\label{sec:offpolicy}

\begin{figure}[t]
    \centering
    \includegraphics[width=\linewidth]{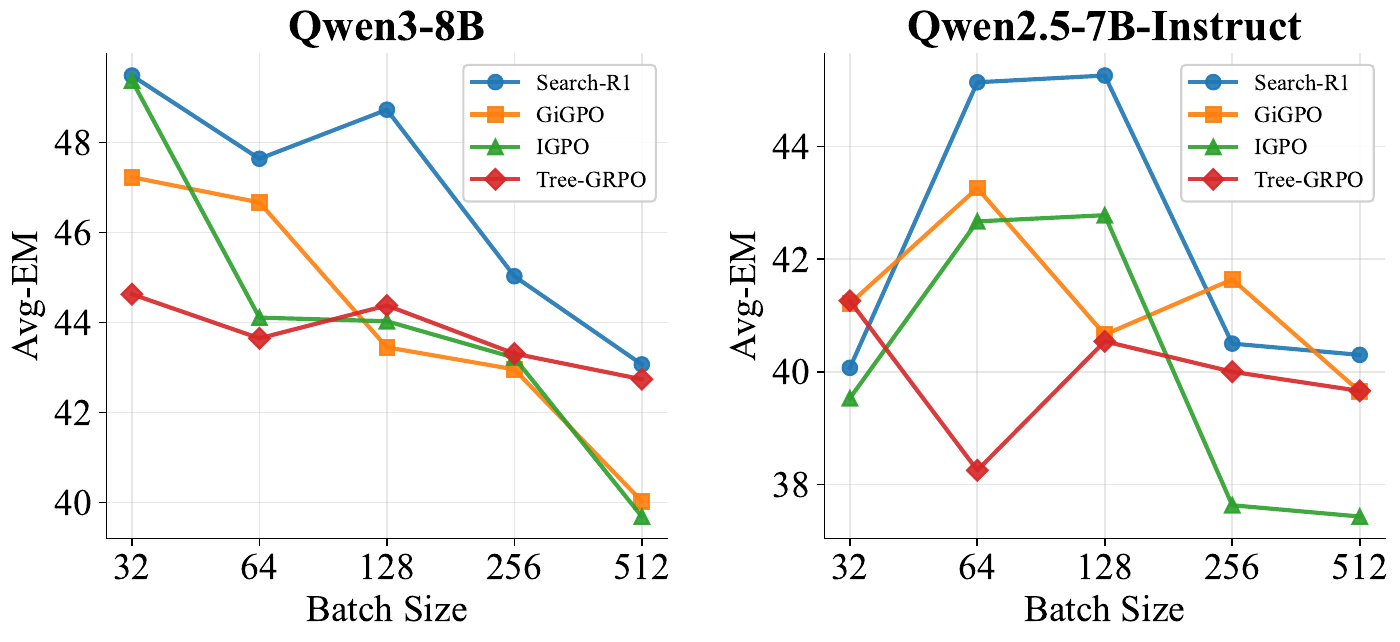}
    \caption{Effect of train batch size on average exact match for four methods.}\label{fig:batchsizes}
\end{figure}

\begin{figure*}[t]
    \centering
    \includegraphics[width=\textwidth]{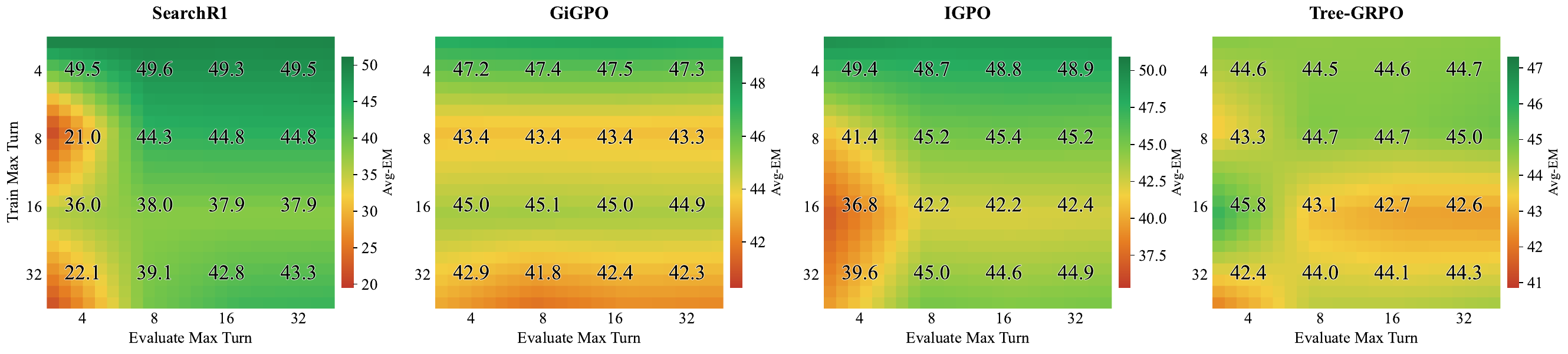}
    \caption{Average EM across five datasets for four methods under varying training (rows) and evaluation (columns) search budgets. All experiments use Qwen3-8B with Hermes format.}\label{fig:max_turns}
\end{figure*}

In GRPO, the model collects a train batch of rollouts and then performs multiple mini-batch updates from it. As the train batch grows larger relative to the mini-batch, later updates operate on trajectories generated by an increasingly outdated policy.
To measure this effect, we fix the mini-batch size at 16 and increase the train batch size exponentially from 32 to 512, training all four methods on Qwen3-8B with the Hermes format, and Qwen2.5-7B-Instruct with the Search-R1 format.

As illustrated in Fig.~\ref{fig:batchsizes}, the two models exhibit distinct trends: for Qwen3-8B, performance generally degrades as batch size increases, with Search-R1 and IGPO showing the steepest drops.
Qwen2.5-7B-Instruct, by contrast, remains relatively stable across batch sizes, with no consistent degradation pattern. 
This indicates that Qwen3-8B is more sensitive to off-policy drift, while Qwen2.5-7B-Instruct tolerates larger batches without systematic performance loss.
The data and off-policy experiments point to a shared underlying factor: both control how fresh the training signal is relative to the current policy. 
Increasing data diversity reduces the chance of learning from repeated, stale examples, while keeping the train batch small ensures that rollouts remain close to the current policy.

\subsubsection{Max Turns}\label{sec:maxturns}

The maximum number of tool-calling turns (i.e., the search budget) is another critical factor affecting model performance.
We conduct experiments on Qwen3-8B (Hermes format) with three algorithms under varying training and evaluation search budgets. 
As shown in Fig.~\ref{fig:max_turns}, none of the four methods achieves more than a 0.3-point gain when the evaluation budget exceeds the training budget. 
Moreover, as the training search budget increases, three methods exhibit performance degradation. 
Comparing the 4-turn train / 4-turn eval setting with the 32-turn train / 32-turn eval setting, Search-R1, GiGPO, and IGPO all score approximately 5\% higher under the 4-turn setting, while Tree-GRPO plateaus around 44\% regardless of budget.
This indicates persistent training instability in long-horizon agent scenarios.

The four methods exhibit distinct patterns under budget mismatch. 
Search-R1 and IGPO degrade noticeably when the evaluation budget falls below 8 turns and is smaller than the training budget, whereas GiGPO and Tree-GRPO remain relatively stable. 
This aligns with their actual search behavior: under a 32-turn budget, Search-R1 and IGPO average approximately 8 search turns, while GiGPO and Tree-GRPO average only 4. 
The former two methods effectively train models to search deeper, but this also makes them more dependent on the available budget. 
In contrast, GiGPO and Tree-GRPO struggle to encourage deeper search behavior, resulting in trajectories that rarely approach the budget limit and thus remain insensitive to evaluation budget changes. 
However, this distinction is also bounded by dataset difficulty: conventional multi-hop datasets can almost always be resolved within 8 searches, leaving limited room to observe the benefits of a deeper search, suggesting that current multi-hop benchmarks offer insufficient complexity to differentiate methods at higher budgets, underscoring the demand for more challenging, long-horizon training data.

\finding{Search budget scaling reveals a ceiling effect: increasing the training and evaluation budget does not yield proportional gains.} %

\section{Summary}

In this paper, we conducted a systematic empirical study of the key factors affecting search agent training: retrieval environment, credit assignment algorithm, and training protocol.
Our findings reveal that the retrieval environment exerts the largest influence on final performance.
Specifically, when the corpus lacks key supporting documents for training questions, the model may earn a reward by guessing from parametric memory rather than through successful retrieval, introducing noise into policy optimization. We further provide practical guidance on training hyperparameters such as batch size and search budget. 
We hope our proposed Wiki-fixed corpus and the accompanying insights establish a stable experimental foundation and reduce the burden of choosing hyperparameters, enabling cleaner methodological comparisons and allowing future work to isolate genuine algorithmic advances more clearly.

\section*{Limitations}

Due to resource constraints, we cannot cover all existing credit assignment algorithms.
Additionally, the high cost of web search APIs prevents us from conducting our full experiment suite (approximately 100 runs) under web retrieval settings; our conclusions are therefore limited to local corpus-based search environments.

\section*{Ethics Statement}
This work was conducted in strict compliance with the ACL Ethics Policy. 
All datasets and models used for the experiment are publicly available, and our usage aligns with their expectations.
For the datasets we used, 2WikiMultihopQA, HotpotQA, and Musique adopt Apache License 2.0, Bamboogle and PopQA adopt MIT License.
For the corpus, Wiki-18 adopts the GNU License.
Furthermore, our work conducted an empirical study of the key factors affecting search agent training.
We do not foresee any negative ethical impacts arising from our work.


\bibliography{custom}

@inproceedings{scalecua,
title={Scale{CUA}: Scaling Open-Source Computer Use Agents with Cross-Platform Data},
author={Zhaoyang Liu and JingJing Xie and Zichen Ding and Zehao Li and Bowen Yang and Zhenyu Wu and Xuehui Wang and Qiushi Sun and Shi Liu and Weiyun Wang and Shenglong Ye and Qingyun Li and Zeyue Tian and Gen Luo and Xiangyu Yue and Biqing Qi and Kai Chen and Bowen Zhou and Yu Qiao and Qifeng Chen and Wenhai Wang},
booktitle={The Fourteenth International Conference on Learning Representations},
year={2026},
url={https://openreview.net/forum?id=yBFUqdJFZn}
}

@article{kimi,
  title={Kimi K2. 5: Visual Agentic Intelligence},
  author={Team, Kimi and Bai, Tongtong and Bai, Yifan and Bao, Yiping and Cai, SH and Cao, Yuan and Charles, Y and Che, HS and Chen, Cheng and Chen, Guanduo and others},
  journal={arXiv preprint arXiv:2602.02276},
  year={2026}
}

@misc{coding-agent,
      title={Scaling Coding Agents via Atomic Skills}, 
      author={Yingwei Ma and Yue Liu and Xinlong Yang and Yanhao Li and Kelin Fu and Yibo Miao and Yuchong Xie and Zhexu Wang and Shing-Chi Cheung},
      year={2026},
      eprint={2604.05013},
      archivePrefix={arXiv},
      primaryClass={cs.SE},
      url={https://arxiv.org/abs/2604.05013}, 
}

@misc{coding-agent-2,
      title={Embarrassingly Simple Self-Distillation Improves Code Generation}, 
      author={Ruixiang Zhang and Richard He Bai and Huangjie Zheng and Navdeep Jaitly and Ronan Collobert and Yizhe Zhang},
      year={2026},
      eprint={2604.01193},
      archivePrefix={arXiv},
      primaryClass={cs.CL},
      url={https://arxiv.org/abs/2604.01193}, 
}

@inproceedings{deep-search-1,
    title = "Adversarial Yet Cooperative: Multi-Perspective Reasoning in Retrieved-Augmented Language Models",
    author = "Xu, Can and
      Yan, Lingyong  and
      Wu, Jiayi  and
      Wang, Haosen  and
      Li, Yuchen and 
      Huang, Jizhou and 
      Yin, Dawei and 
      Li, Xiang",
    booktitle = "Findings of the Association for Computational Linguistics: ACL 2026",
    month = jul,
    year = "2026",
    address = "San Diego, USA",
    publisher = "Association for Computational Linguistics",
}

@misc{deep-search-2,
      title={DR Tulu: Reinforcement Learning with Evolving Rubrics for Deep Research}, 
      author={Rulin Shao and Akari Asai and Shannon Zejiang Shen and Hamish Ivison and Varsha Kishore and Jingming Zhuo and Xinran Zhao and Molly Park and Samuel G. Finlayson and David Sontag and Tyler Murray and Sewon Min and Pradeep Dasigi and Luca Soldaini and Faeze Brahman and Wen-tau Yih and Tongshuang Wu and Luke Zettlemoyer and Yoon Kim and Hannaneh Hajishirzi and Pang Wei Koh},
      year={2025},
      eprint={2511.19399},
      archivePrefix={arXiv},
      primaryClass={cs.CL},
      url={https://arxiv.org/abs/2511.19399}, 
}

@misc{Qwen3,
      title={Qwen3 Technical Report}, 
      author={An Yang and Anfeng Li and Baosong Yang and Beichen Zhang and Binyuan Hui and Bo Zheng and Bowen Yu and Chang Gao and Chengen Huang and Chenxu Lv and Chujie Zheng and Dayiheng Liu and Fan Zhou and Fei Huang and Feng Hu and Hao Ge and Haoran Wei and Huan Lin and Jialong Tang and Jian Yang and Jianhong Tu and Jianwei Zhang and Jianxin Yang and Jiaxi Yang and Jing Zhou and Jingren Zhou and Junyang Lin and Kai Dang and Keqin Bao and Kexin Yang and Le Yu and Lianghao Deng and Mei Li and Mingfeng Xue and Mingze Li and Pei Zhang and Peng Wang and Qin Zhu and Rui Men and Ruize Gao and Shixuan Liu and Shuang Luo and Tianhao Li and Tianyi Tang and Wenbiao Yin and Xingzhang Ren and Xinyu Wang and Xinyu Zhang and Xuancheng Ren and Yang Fan and Yang Su and Yichang Zhang and Yinger Zhang and Yu Wan and Yuqiong Liu and Zekun Wang and Zeyu Cui and Zhenru Zhang and Zhipeng Zhou and Zihan Qiu},
      year={2025},
      eprint={2505.09388},
      archivePrefix={arXiv},
      primaryClass={cs.CL},
      url={https://arxiv.org/abs/2505.09388}, 
}

@inproceedings{flashattention,
title={FlashAttention-2: Faster Attention with Better Parallelism and Work Partitioning},
author={Tri Dao},
booktitle={The Twelfth International Conference on Learning Representations},
year={2024},
url={https://openreview.net/forum?id=mZn2Xyh9Ec}
}

@inproceedings{vllm,
author = {Kwon, Woosuk and Li, Zhuohan and Zhuang, Siyuan and Sheng, Ying and Zheng, Lianmin and Yu, Cody Hao and Gonzalez, Joseph and Zhang, Hao and Stoica, Ion},
title = {Efficient Memory Management for Large Language Model Serving with PagedAttention},
year = {2023},
isbn = {9798400702297},
publisher = {Association for Computing Machinery},
address = {New York, NY, USA},
url = {https://doi.org/10.1145/3600006.3613165},
doi = {10.1145/3600006.3613165},
booktitle = {Proceedings of the 29th Symposium on Operating Systems Principles},
pages = {611–626},
numpages = {16},
location = {Koblenz, Germany},
series = {SOSP '23}
}

@inproceedings{translation-1,
author = {Xu, Haoran and Sharaf, Amr and Chen, Yunmo and Tan, Weiting and Shen, Lingfeng and Van Durme, Benjamin and Murray, Kenton and Kim, Young Jin},
title = {Contrastive preference optimization: pushing the boundaries of LLM performance in machine translation},
year = {2024},
publisher = {JMLR.org},
booktitle = {Proceedings of the 41st International Conference on Machine Learning},
articleno = {2275},
numpages = {21},
location = {Vienna, Austria},
series = {ICML'24}
}

@inproceedings{translation-2,
    title = "{MT}-R1-Zero: Advancing {LLM}-based Machine Translation via R1-Zero-like Reinforcement Learning",
    author = "Feng, Zhaopeng  and
      Cao, Shaosheng  and
      Ren, Jiahan  and
      Su, Jiayuan  and
      Chen, Ruizhe  and
      Zhang, Yan  and
      Wu, Jian  and
      Liu, Zuozhu",
    editor = "Christodoulopoulos, Christos  and
      Chakraborty, Tanmoy  and
      Rose, Carolyn  and
      Peng, Violet",
    booktitle = "Findings of the Association for Computational Linguistics: EMNLP 2025",
    month = nov,
    year = "2025",
    address = "Suzhou, China",
    publisher = "Association for Computational Linguistics",
    url = "https://aclanthology.org/2025.findings-emnlp.1015/",
    doi = "10.18653/v1/2025.findings-emnlp.1015",
    pages = "18685--18702",
    ISBN = "979-8-89176-335-7",
}

@inproceedings{translation-3,
  title={Asymmetric conflict and synergy in post-training for llm-based multilingual machine translation},
  author={Zheng, Tong and Wen, Yan and Bao, Huiwen and Guo, Junfeng and Huang, Heng},
  booktitle={Findings of the Association for Computational Linguistics: ACL 2025},
  pages={18362--18383},
  year={2025}
}

@misc{reasoning-1,
      title={DeepSeekMath: Pushing the Limits of Mathematical Reasoning in Open Language Models}, 
      author={Zhihong Shao and Peiyi Wang and Qihao Zhu and Runxin Xu and Junxiao Song and Xiao Bi and Haowei Zhang and Mingchuan Zhang and Y. K. Li and Y. Wu and Daya Guo},
      year={2024},
      eprint={2402.03300},
      archivePrefix={arXiv},
      primaryClass={cs.CL},
      url={https://arxiv.org/abs/2402.03300}, 
}

@inproceedings{reasoning-2,
title={Chain of Thought Prompting Elicits Reasoning in Large Language Models},
author={Jason Wei and Xuezhi Wang and Dale Schuurmans and Maarten Bosma and brian ichter and Fei Xia and Ed H. Chi and Quoc V Le and Denny Zhou},
booktitle={Advances in Neural Information Processing Systems},
editor={Alice H. Oh and Alekh Agarwal and Danielle Belgrave and Kyunghyun Cho},
year={2022},
url={https://openreview.net/forum?id=_VjQlMeSB_J}
}

@misc{reasoning-3,
      title={DeepSeekMath-V2: Towards Self-Verifiable Mathematical Reasoning}, 
      author={Zhihong Shao and Yuxiang Luo and Chengda Lu and Z. Z. Ren and Jiewen Hu and Tian Ye and Zhibin Gou and Shirong Ma and Xiaokang Zhang},
      year={2025},
      eprint={2511.22570},
      archivePrefix={arXiv},
      primaryClass={cs.AI},
      url={https://arxiv.org/abs/2511.22570}, 
}

@misc{chatbot2agent,
      title={From LLM Reasoning to Autonomous AI Agents: A Comprehensive Review}, 
      author={Mohamed Amine Ferrag and Norbert Tihanyi and Merouane Debbah},
      year={2026},
      eprint={2504.19678},
      archivePrefix={arXiv},
      primaryClass={cs.AI},
      url={https://arxiv.org/abs/2504.19678}, 
}

@inproceedings{writing-1,
    title = "Igniting Creative Writing in Small Language Models: {LLM}-as-a-Judge versus Multi-Agent Refined Rewards",
    author = "Wei, Xiaolong  and
      Lu, Bo  and
      Zhang, Xingyu  and
      Zhao, Zhejun  and
      Shen, Dongdong  and
      Xia, Long  and
      Yin, Dawei",
    editor = "Christodoulopoulos, Christos  and
      Chakraborty, Tanmoy  and
      Rose, Carolyn  and
      Peng, Violet",
    booktitle = "Proceedings of the 2025 Conference on Empirical Methods in Natural Language Processing",
    month = nov,
    year = "2025",
    address = "Suzhou, China",
    publisher = "Association for Computational Linguistics",
    url = "https://aclanthology.org/2025.emnlp-main.868/",
    doi = "10.18653/v1/2025.emnlp-main.868",
    pages = "17160--17186",
    ISBN = "979-8-89176-332-6",
}

@inproceedings{writing-2,
author = {Qin, Hua Xuan and Jin, Shan and Gao, Ze and Fan, Mingming and Hui, Pan},
title = {CharacterMeet: Supporting Creative Writers' Entire Story Character Construction Processes Through Conversation with LLM-Powered Chatbot Avatars},
year = {2024},
isbn = {9798400703300},
publisher = {Association for Computing Machinery},
address = {New York, NY, USA},
url = {https://doi.org/10.1145/3613904.3642105},
doi = {10.1145/3613904.3642105},
booktitle = {Proceedings of the 2024 CHI Conference on Human Factors in Computing Systems},
articleno = {1051},
numpages = {19},
location = {Honolulu, HI, USA},
series = {CHI '24}
}

@inproceedings{writing-3,
title={Modifying Large Language Model Post-Training for Diverse Creative Writing},
author={John Joon Young Chung and Vishakh Padmakumar and Melissa Roemmele and Yuqian Sun and Max Kreminski},
booktitle={Second Conference on Language Modeling},
year={2025},
url={https://openreview.net/forum?id=1Pmuw08LoM}
}

@inproceedings{swebench,
title={{SWE}-bench: Can Language Models Resolve Real-world Github Issues?},
author={Carlos E Jimenez and John Yang and Alexander Wettig and Shunyu Yao and Kexin Pei and Ofir Press and Karthik R Narasimhan},
booktitle={The Twelfth International Conference on Learning Representations},
year={2024},
url={https://openreview.net/forum?id=VTF8yNQM66}
}

@inproceedings{osworld,
title={{OSW}orld: Benchmarking Multimodal Agents for Open-Ended Tasks in Real Computer Environments},
author={Tianbao Xie and Danyang Zhang and Jixuan Chen and Xiaochuan Li and Siheng Zhao and Ruisheng Cao and Toh Jing Hua and Zhoujun Cheng and Dongchan Shin and Fangyu Lei and Yitao Liu and Yiheng Xu and Shuyan Zhou and Silvio Savarese and Caiming Xiong and Victor Zhong and Tao Yu},
booktitle={The Thirty-eight Conference on Neural Information Processing Systems Datasets and Benchmarks Track},
year={2024},
url={https://openreview.net/forum?id=tN61DTr4Ed}
}

@misc{GAIA,
      title={GAIA: a benchmark for General AI Assistants}, 
      author={Grégoire Mialon and Clémentine Fourrier and Craig Swift and Thomas Wolf and Yann LeCun and Thomas Scialom},
      year={2023},
      eprint={2311.12983},
      archivePrefix={arXiv},
      primaryClass={cs.CL},
      url={https://arxiv.org/abs/2311.12983}, 
}

@misc{browsecomp,
      title={BrowseComp: A Simple Yet Challenging Benchmark for Browsing Agents}, 
      author={Jason Wei and Zhiqing Sun and Spencer Papay and Scott McKinney and Jeffrey Han and Isa Fulford and Hyung Won Chung and Alex Tachard Passos and William Fedus and Amelia Glaese},
      year={2025},
      eprint={2504.12516},
      archivePrefix={arXiv},
      primaryClass={cs.CL},
      url={https://arxiv.org/abs/2504.12516}, 
}

@inproceedings{hotpotqa,
    title = "{H}otpot{QA}: A Dataset for Diverse, Explainable Multi-hop Question Answering",
    author = "Yang, Zhilin  and
      Qi, Peng  and
      Zhang, Saizheng  and
      Bengio, Yoshua  and
      Cohen, William  and
      Salakhutdinov, Ruslan  and
      Manning, Christopher D.",
    editor = "Riloff, Ellen  and
      Chiang, David  and
      Hockenmaier, Julia  and
      Tsujii, Jun{'}ichi",
    booktitle = "Proceedings of the 2018 Conference on Empirical Methods in Natural Language Processing",
    month = oct # "-" # nov,
    year = "2018",
    address = "Brussels, Belgium",
    publisher = "Association for Computational Linguistics",
    url = "https://aclanthology.org/D18-1259/",
    doi = "10.18653/v1/D18-1259",
    pages = "2369--2380",
}

@inproceedings{2wiki,
    title = "Constructing A Multi-hop {QA} Dataset for Comprehensive Evaluation of Reasoning Steps",
    author = "Ho, Xanh  and
      Duong Nguyen, Anh-Khoa  and
      Sugawara, Saku  and
      Aizawa, Akiko",
    editor = "Scott, Donia  and
      Bel, Nuria  and
      Zong, Chengqing",
    booktitle = "Proceedings of the 28th International Conference on Computational Linguistics",
    month = dec,
    year = "2020",
    address = "Barcelona, Spain (Online)",
    publisher = "International Committee on Computational Linguistics",
    url = "https://aclanthology.org/2020.coling-main.580/",
    doi = "10.18653/v1/2020.coling-main.580",
    pages = "6609--6625",
}

@article{musique,
    title = "{M}u{S}i{Q}ue: Multihop Questions via Single-hop Question Composition",
    author = "Trivedi, Harsh  and
      Balasubramanian, Niranjan  and
      Khot, Tushar  and
      Sabharwal, Ashish",
    editor = "Roark, Brian  and
      Nenkova, Ani",
    journal = "Transactions of the Association for Computational Linguistics",
    volume = "10",
    year = "2022",
    address = "Cambridge, MA",
    publisher = "MIT Press",
    url = "https://aclanthology.org/2022.tacl-1.31/",
    doi = "10.1162/tacl_a_00475",
    pages = "539--554",
}

@inproceedings{bamboogle,
    title = "Measuring and Narrowing the Compositionality Gap in Language Models",
    author = "Press, Ofir  and
      Zhang, Muru  and
      Min, Sewon  and
      Schmidt, Ludwig  and
      Smith, Noah  and
      Lewis, Mike",
    editor = "Bouamor, Houda  and
      Pino, Juan  and
      Bali, Kalika",
    booktitle = "Findings of the Association for Computational Linguistics: EMNLP 2023",
    month = dec,
    year = "2023",
    address = "Singapore",
    publisher = "Association for Computational Linguistics",
    url = "https://aclanthology.org/2023.findings-emnlp.378/",
    doi = "10.18653/v1/2023.findings-emnlp.378",
    pages = "5687--5711",
}

@misc{arlarena,
      title={ARLArena: A Unified Framework for Stable Agentic Reinforcement Learning}, 
      author={Xiaoxuan Wang and Han Zhang and Haixin Wang and Yidan Shi and Ruoyan Li and Kaiqiao Han and Chenyi Tong and Haoran Deng and Renliang Sun and Alexander Taylor and Yanqiao Zhu and Jason Cong and Yizhou Sun and Wei Wang},
      year={2026},
      eprint={2602.21534},
      archivePrefix={arXiv},
      primaryClass={cs.AI},
      url={https://arxiv.org/abs/2602.21534}, 
}

@misc{ragen2,
      title={RAGEN-2: Reasoning Collapse in Agentic RL}, 
      author={Zihan Wang and Chi Gui and Xing Jin and Qineng Wang and Licheng Liu and Kangrui Wang and Shiqi Chen and Linjie Li and Zhengyuan Yang and Pingyue Zhang and Yiping Lu and Jiajun Wu and Li Fei-Fei and Lijuan Wang and Yejin Choi and Manling Li},
      year={2026},
      eprint={2604.06268},
      archivePrefix={arXiv},
      primaryClass={cs.LG},
      url={https://arxiv.org/abs/2604.06268}, 
}

@misc{llds,
      title={On Group Relative Policy Optimization Collapse in Agent Search: The Lazy Likelihood-Displacement}, 
      author={Wenlong Deng and Yushu Li and Boying Gong and Yi Ren and Christos Thrampoulidis and Xiaoxiao Li},
      year={2026},
      eprint={2512.04220},
      archivePrefix={arXiv},
      primaryClass={cs.CL},
      url={https://arxiv.org/abs/2512.04220}, 
}

@inproceedings{wiki18,
    title = "Dense Passage Retrieval for Open-Domain Question Answering",
    author = "Karpukhin, Vladimir  and
      Oguz, Barlas  and
      Min, Sewon  and
      Lewis, Patrick  and
      Wu, Ledell  and
      Edunov, Sergey  and
      Chen, Danqi  and
      Yih, Wen-tau",
    editor = "Webber, Bonnie  and
      Cohn, Trevor  and
      He, Yulan  and
      Liu, Yang",
    booktitle = "Proceedings of the 2020 Conference on Empirical Methods in Natural Language Processing (EMNLP)",
    month = nov,
    year = "2020",
    address = "Online",
    publisher = "Association for Computational Linguistics",
    url = "https://aclanthology.org/2020.emnlp-main.550/",
    doi = "10.18653/v1/2020.emnlp-main.550",
    pages = "6769--6781",
}

@inproceedings{searchr1,
title={Search-R1: Training {LLM}s to Reason and Leverage Search Engines with Reinforcement Learning},
author={Bowen Jin and Hansi Zeng and Zhenrui Yue and Jinsung Yoon and Sercan O Arik and Dong Wang and Hamed Zamani and Jiawei Han},
booktitle={Second Conference on Language Modeling},
year={2025},
url={https://openreview.net/forum?id=Rwhi91ideu}
}

@misc{r1searcher,
      title={R1-Searcher: Incentivizing the Search Capability in LLMs via Reinforcement Learning}, 
      author={Huatong Song and Jinhao Jiang and Yingqian Min and Jie Chen and Zhipeng Chen and Wayne Xin Zhao and Lei Fang and Ji-Rong Wen},
      year={2025},
      eprint={2503.05592},
      archivePrefix={arXiv},
      primaryClass={cs.AI},
      url={https://arxiv.org/abs/2503.05592}, 
}

@inproceedings{reasonrag,
title={Process vs. Outcome Reward: Which is Better for Agentic {RAG} Reinforcement Learning},
author={Wenlin Zhang and Xiangyang Li and Kuicai Dong and Yichao Wang and Pengyue Jia and Xiaopeng Li and Yingyi Zhang and Derong Xu and Zhaocheng Du and Huifeng Guo and Ruiming Tang and Xiangyu Zhao},
booktitle={The Thirty-ninth Annual Conference on Neural Information Processing Systems},
year={2025},
url={https://openreview.net/forum?id=h3LlJ6Bh4S}
}

@inproceedings{gigpo,
title={Group-in-Group Policy Optimization for {LLM} Agent Training},
author={Lang Feng and Zhenghai Xue and Tingcong Liu and Bo An},
booktitle={The Thirty-ninth Annual Conference on Neural Information Processing Systems},
year={2025},
url={https://openreview.net/forum?id=QXEhBMNrCW}
}

@inproceedings{igpo,
title={Information Gain-based Policy Optimization: A Simple and Effective Approach for Multi-Turn Search Agents},
author={Guoqing Wang and Sunhao Dai and Guangze Ye and Zeyu Gan and Wei Yao and Yong Deng and Xiaofeng Wu and Zhenzhe Ying},
booktitle={The Fourteenth International Conference on Learning Representations},
year={2026},
url={https://openreview.net/forum?id=qkWP6phrvZ}
}

@article{tree-grpo,
  title={Tree Search for LLM Agent Reinforcement Learning}, 
  author={Yuxiang Ji and Ziyu Ma and Yong Wang and Guanhua Chen and Xiangxiang Chu and Liaoni Wu},
  journal={arXiv preprint arXiv:2509.21240},
  year={2025}
}

@inproceedings{stepsearch,
    title = "{S}tep{S}earch: Igniting {LLM}s Search Ability via Step-Wise Proximal Policy Optimization",
    author = "Zheng, Xuhui  and
      An, Kang  and
      Wang, Ziliang  and
      Wang, Yuhang  and
      Wu, Yichao",
    editor = "Christodoulopoulos, Christos  and
      Chakraborty, Tanmoy  and
      Rose, Carolyn  and
      Peng, Violet",
    booktitle = "Proceedings of the 2025 Conference on Empirical Methods in Natural Language Processing",
    month = nov,
    year = "2025",
    address = "Suzhou, China",
    publisher = "Association for Computational Linguistics",
    url = "https://aclanthology.org/2025.emnlp-main.1106/",
    doi = "10.18653/v1/2025.emnlp-main.1106",
    pages = "21805--21830",
    ISBN = "979-8-89176-332-6"
}

@misc{dapo,
      title={DAPO: An Open-Source LLM Reinforcement Learning System at Scale}, 
      author={Qiying Yu and Zheng Zhang and Ruofei Zhu and Yufeng Yuan and Xiaochen Zuo and Yu Yue and Weinan Dai and Tiantian Fan and Gaohong Liu and Lingjun Liu and Xin Liu and Haibin Lin and Zhiqi Lin and Bole Ma and Guangming Sheng and Yuxuan Tong and Chi Zhang and Mofan Zhang and Wang Zhang and Hang Zhu and Jinhua Zhu and Jiaze Chen and Jiangjie Chen and Chengyi Wang and Hongli Yu and Yuxuan Song and Xiangpeng Wei and Hao Zhou and Jingjing Liu and Wei-Ying Ma and Ya-Qin Zhang and Lin Yan and Mu Qiao and Yonghui Wu and Mingxuan Wang},
      year={2025},
      eprint={2503.14476},
      archivePrefix={arXiv},
      primaryClass={cs.LG},
      url={https://arxiv.org/abs/2503.14476}, 
}

@inproceedings{popqa,
    title = "When Not to Trust Language Models: Investigating Effectiveness of Parametric and Non-Parametric Memories",
    author = "Mallen, Alex  and
      Asai, Akari  and
      Zhong, Victor  and
      Das, Rajarshi  and
      Khashabi, Daniel  and
      Hajishirzi, Hannaneh",
    editor = "Rogers, Anna  and
      Boyd-Graber, Jordan  and
      Okazaki, Naoaki",
    booktitle = "Proceedings of the 61st Annual Meeting of the Association for Computational Linguistics (Volume 1: Long Papers)",
    month = jul,
    year = "2023",
    address = "Toronto, Canada",
    publisher = "Association for Computational Linguistics",
    url = "https://aclanthology.org/2023.acl-long.546/",
    doi = "10.18653/v1/2023.acl-long.546",
    pages = "9802--9822"
}

@article{verl,
  title   = {HybridFlow: A Flexible and Efficient RLHF Framework},
  author  = {Guangming Sheng and Chi Zhang and Zilingfeng Ye and Xibin Wu and Wang Zhang and Ru Zhang and Yanghua Peng and Haibin Lin and Chuan Wu},
  year    = {2024},
  journal = {arXiv preprint arXiv: 2409.19256}
}

@misc{hermes,
      title={Hermes 3 Technical Report}, 
      author={Ryan Teknium and Jeffrey Quesnelle and Chen Guang},
      year={2024},
      eprint={2408.11857},
      archivePrefix={arXiv},
      primaryClass={cs.CL},
      url={https://arxiv.org/abs/2408.11857}, 
}

@misc{calibadv,
      title={Negative Advantage Is a Double-Edged Sword: Calibrating Advantage in GRPO for Deep Search}, 
      author={Jiayi Wu and Ruobing Xie and Zeqian Huang and Lei Jiang and Can Xu and Kangyang Luo and Ming Gao and Xiang Li},
      year={2026},
      eprint={2604.18235},
      archivePrefix={arXiv},
      primaryClass={cs.CL},
      url={https://arxiv.org/abs/2604.18235}, 
}

@misc{cua2025,
title={Computer-Using Agent: Introducing a universal interface for AI to interact with the digital world},
author={OpenAI},
year={2025},
url={https://openai.com/index/computer-using-agent},
}

@article{yang2026symphony,
  title={Os-symphony: A holistic framework for robust and generalist computer-using agent},
  author={Yang, Bowen and Jin, Kaiming and Wu, Zhenyu and Liu, Zhaoyang and Sun, Qiushi and Li, Zehao and Xie, JingJing and Liu, Zhoumianze and Xu, Fangzhi and Cheng, Kanzhi and others},
  journal={arXiv preprint arXiv:2601.07779},
  year={2026}
}

@software{SwanLab,
  author = {Zeyi Lin and Shaohong Chen and Kang Li and Qiushan Jiang and Zirui Cai and Kaifang Ji and {The SwanLab team}},
  doi = {10.5281/zenodo.11100550},
  license = {Apache-2.0},
  title = {{SwanLab}},
  url = {https://github.com/swanhubx/swanlab},
  year = {2023}
}
\clearpage
\appendix

\section*{Appendix}
\label{sec:appendix}

\section{Detailed Algorithms}\label{app:algorithm}

\subsection{Preliminaries}

We model the search agent as a Partially Observable Markov Decision Process (POMDP), defined by the tuple $(\mathcal S,\mathcal A,\mathcal O,\mathcal T,\mathcal R)$.

\textbf{State space $\mathcal S$.} The environment state is a fixed retrieval corpus $\mathcal D= \{d_1,d_2,\cdots,d_N\}$, which the agent can only access through retrieval actions.

\textbf{Action space $\mathcal A$.} At each step $t$, the agent produces an action $a_t = (\text{think}_t,\text{query}_t)$: a \textit{thinking} step in which the agent reasons about what information is still needed, followed by a \textit{tool call} that issues a search query to the retrieval environment.

\textbf{Observation space $\mathcal O$.} After executing action $a_t$, the agent receives an observation $o_t$, the set of passages returned by the retrieval system in response to $\text{query}_t$. These observations accumulate into the interaction history: $h_t = (a_1, o_1, \ldots, a_{t-1}, o_{t-1})$.

\textbf{Transition $\mathcal T$.} Since the corpus $\mathcal D$ is static, the environment transition is fully governed by the retrieval function:
$o_t=\text{Retrieve}(\text{query}_t;\mathcal D).$

\textbf{Reward $\mathcal R$.} A reward function $r(h_T,a_T)$ assigns a scalar score at the end of the trajectory. The design of this reward, outcome-based versus process-based, is a central variable we investigate.

The agent's policy $\pi_\theta(a_t\mid h_t)$ is parameterized by an LLM with parameters $\theta$, conditioned on the interaction history $h$ defined above. We denote the ground-truth answer to the input question as $\text{gt}$.

\subsection{Training Algorithms}



\textbf{Search-R1.} Search-R1 uses a simple outcome-based reward with exact match (EM) verification. The reward function assigns $r=1$ if the extracted answer is correct and properly formatted, $r=0.25$ if the answer is correct but the output contains an excessive number of answer tags, and $r=0$ otherwise. The advantage is computed at the trajectory level and assigned uniformly to all tokens:
\begin{equation}
    \hat{A}^{(i)} = \frac{r_i-\text{mean}(r^{(j)})_{j=1}^G}{\text{std}(r^{(j)})_{j=1}^G}.\label{eq2}
\end{equation}

\textbf{GiGPO.} GiGPO~\cite{gigpo} constructs a two-level advantage: episode-level and step-level. The episode-level advantage follows Eq.~\ref{eq2}. For step-level credit assignment, GiGPO groups actions by anchor states: environment states matched via textual similarity rather than step index. 
Actions sharing the same anchor state are grouped regardless of which trajectory they belong to, and advantages are normalized within each group. The final advantage is: $\hat A^{(i)} = \hat A^{(i)\text{E}} + \omega \hat A^{(i)\text{S}}$, where $\omega$ controls the relative weight of step-level credit, and $\hat A^{(i)\text{S}}$ is defined as:
\begin{equation}
\small
    \hat A^{(i)\text{S}}\left(a_t^{(i)}\right) = \frac{r_t^{(i)} - \text{mean}\left(\left\{r_t \mid (a_t, r_t) \in G^S(\tilde s)\right\}\right)}{\text{std}\left(\left\{r_t \mid \left(a_t, r_t\right) \in G^S(\tilde s) \right\}\right)}.
\end{equation}
Here, $G^S(\tilde s)$ denotes the set of action-reward pairs grouped under the same anchor state $\tilde s$.

\textbf{IGPO.} IGPO~\cite{igpo} models the multi-turn agent-environment interaction as an incremental process of acquiring information toward the ground truth. 
It introduces an intrinsic reward based on information gain: at each turn, the model computes the probability of generating the ground truth, and the turn-level reward is defined as the log-probability difference between consecutive turns. 
Formally, the turn-level reward is:
\begin{equation}
    r_t = \log \pi_\theta(\text{gt}\mid h_t,a_t) - \log \pi_\theta(\text{gt}\mid h_{t-1},a_{t-1}).
\end{equation}
The outcome reward at the final turn uses F1 overlap when the output format is valid and a fixed penalty $\lambda_\text{Format}$ otherwise:
\begin{equation}
r_T =
\begin{cases}
\text{F1}(a_T, \text{gt}), & \text{if the output format is valid},\\
\lambda_{\text{Format}}, & \text{otherwise}.
\end{cases}
\end{equation}
Then, IGPO normalizes intermediate and outcome rewards separately via group statistics:
\begin{equation}
\tilde r_t^{(i)} =
\begin{cases}
\dfrac{r_t^{(i)} - \mu_{1:T-1}}{\sigma_{1:T-1}}, & 1 \le t \le T-1,\\[6pt]
\dfrac{r_t^{(i)} - \mu_T}{\sigma_T}, & t = T,
\end{cases}
\end{equation}
where $\mu_{1:T-1} = \mathrm{mean}\bigl(\{r_{t'}^{(j)}\}_{j=1,t'=1}^{G,\;T-1}\bigr),$ $\sigma_{1:T-1} = \mathrm{std}\bigl(\{r_{t'}^{(j)}\}_{j=1,t'=1}^{G,\;T-1}\bigr).$ The final advantage is defined as a discounted sum:
\begin{equation}
    \hat A^{(i)}_t = \sum_{k=t}^T \gamma^{k-t} \tilde r_k^{(i)}.
\end{equation}

\textbf{Tree-GRPO.} Tree-GRPO adopts a sample-then-expand paradigm: at each iteration, it randomly selects intermediate nodes from existing trajectories and rolls out new branches from these states. 
After several expansion iterations, this yields a set of trees whose branches share common prefixes. 
The outcome reward is then computed for each complete branch and assigned to its constituent steps. 
Specifically, Tree-GRPO estimates grouped advantages at two levels. 
The intra-tree advantage normalizes rewards among trajectories within the same tree $\mathcal T_i$:
\begin{equation}
    \hat{A}^{(i)}_{\text{intra}} = \frac{r^{(i)} - \text{mean}\left(\{r^{(j)}\}_{j \in \mathcal{T}_i}\right)}{\text{std}\left(\{r^{(j)}\}_{j \in \mathcal{T}_i}\right)}.
\end{equation}
Since the limited number of branches within each tree may lead to unreliable baseline estimation, Tree-GRPO also computes an inter-tree advantage across all trajectories in the group:
\begin{equation}
    \hat{A}^{(i)}_{\text{inter}} = \frac{r^{(i)} - \text{mean}(\{r^{(j)}\}_{j=1}^G)}{\text{std}(\{r^{(j)}\}_{j=1}^G)}.
\end{equation}
The final advantage combines both levels: 
\begin{equation}
    \hat{A}^{(i)} = \hat{A}^{(i)}_{\text{intra}} + \hat{A}^{(i)}_{\text{inter}}.
\end{equation}

\section{Detailed Hyper-parameters}\label{app:hyper}

We provide a detailed table including the training and evaluating parameters in Tab.~\ref{tab:hyperparameters}. We conducted all the experiments with VeRL 0.8.0.dev0 version. We will also open-source all the training curves on the Swanlab~\cite{SwanLab}, which provides every hyperparameter in the corresponding training cards. For our computational resource, we leverage 8$\times$ NVIDIA H200 GPUs to serve the retrieval corpus, and another 8$\times$ NVIDIA H200 GPUs to train the policy model for each experiment. For each experiment, it takes 3-50 hours, depending on the training setting.

\begin{table}[tbp]
    \centering
    \resizebox{!}{0.35\linewidth}{\begin{tabular}{c|c c}
    \toprule
    Parameter & Training & Inference \\
    \midrule
        rollout\_n & 8 & 4 \\
        clip\_low & 0.2 & N/A\\
        clip\_high & 0.28 & N/A\\
        clip\_ratio\_c & 3 & N/A\\
        learning\_rate & $5e-6^\star$ & N/A\\
        training\_epochs & 1 & N/A \\
        warmup\_steps & 0 & N/A\\
        max\_response\_length & 4096 & 4096 \\
        train\_prompt\_batchsize & 32 & N/A\\
        train\_prompt\_mini\_batchsize & 16 & N/A\\
        temperature & 1.0 & 0.6\\
        top\_p & 1.0 & 1.0\\
    \bottomrule
    \end{tabular}}
    \caption{Main hyperparameters for our experiments. $\star$ means this value varies among different training algorithms to avoid training collapse.}
    \label{tab:hyperparameters}
\end{table}

\section{Prompt Template}\label{app:prompt}

For our training and inference, we adopt the same zero-shot prompt as shown in Tab.~\ref{tab:Hermes}, Tab.~\ref{tab:Search-R1}, for hermes format and Search-R1 format, respectively.

\begin{table}[htb]
\centering
\phantomsection
\begin{tcolorbox}[colback=white!95!gray,colframe=gray!50!black,rounded corners,label={scale-depression}, title={Our Prompt}]
{\small system

You are a helpful assistant specialized in information retrieval.

You will be given a user query and you are required to answer the question based on the information retrieved.

\# Rules

- You may call functions multiple times across turns to complete the task. Each turn, you MUST output exactly ONE function call before waiting for the result.

- Base your answer strictly on the information retrieved from the function calls.

- After gathering all necessary information, provide your final answer within <answer></answer> XML tags.

\# Response Format

1. If you need to call a function, output the function call in the specified XML format.

2. When you have enough information to answer, provide your final response as:

<answer>

[Your concise answer here]

</answer>

\# Important

- Do NOT make up information that is not present in the retrieved documents.

- Be concise but complete in your final answer.

\# Tools

You may call one or more functions to assist with the user query.

You are provided with function signatures within <tools></tools> XML tags:

<tools>

\{"type": "function", "function": \{"name": "local\_search", "description": "Search a local corpus via embedding service and return top-k documents.", "parameters": \{"type": "object", "properties": \{"query\_list": \{"type": "array", "description": "A list of fully-formed semantic queries. The tool will return search results for each query."\}, "k": \{"type": "integer", "description": "Top-k documents to return; int or list[int] aligned to query\_list. If k is an integer, the tool will return the top-k documents for all queries."\}\}, "required": ["query\_list"]\}\}\}

</tools>

For each function call, return a json object with function name and arguments within <tool\_call></tool\_call> XML tags:

<tool\_call>

{"name": <function-name>, "arguments": <args-json-object>}

</tool\_call>

user

Answer the question. If you need more context, call the function `local\_search` to search the local HotpotQA corpus. Wrap the final answer inside <answer>...</answer>.

Question: \{QUESTION\}

assistant}

\end{tcolorbox}
\caption{Prompt template, Hermes Format.}
\label{tab:Hermes}
\end{table}

\begin{table}[htb]
\centering
\phantomsection
\begin{tcolorbox}[colback=white!95!gray,colframe=gray!50!black,rounded corners,label={scale-depression}, title={Our Prompt}]
{\small system

You are a helpful assistant specialized in information retrieval.

You will be given a user query, and you are required to answer the question based on the information retrieved.

\# Tools

You have access to a search engine. To search, wrap your query in <search> and </search> tags, like this:

<search>your search query</search>

The search results will be returned inside <information>...</information> tags.

\# Rules

- You may search multiple times to gather information. Each turn, output exactly ONE <search>...</search> tag before waiting for results.

- Base your answer strictly on the information retrieved from searches.

- After gathering all necessary information, provide your final answer within <answer></answer> XML tags.

\# Response Format

1. If you need to search, output: <search>your query here</search>

2. When you have enough information, provide your final response as:

<answer>

[Your concise answer here]

</answer>

\# Important

- Do NOT make up information that is not present in the retrieved documents.

- Be concise but complete in your final answer.

user

Answer the question based on the search results. Use <search>query</search> to search for information. Wrap the final answer inside <answer>...</answer>.

Question: \{QUESTION\}

assistant}

\end{tcolorbox}
\caption{Prompt template, Search-R1 Format.}
\label{tab:Search-R1}
\end{table}

\section{Detailed Results}\label{app:detailed_res}

Following the structure in the main text, we report the detailed metrics for each dataset here.

Corresponding to Fig.~\ref{fig:env}, we report the detailed results in Tab.~\ref{tab:evn_detailed}.

\begin{table*}[htbp]
\centering
\resizebox{\textwidth}{!}{\begin{tabular}{cccccccccc}
\toprule
Method& Train Env. & Test Env. & 2Wiki & Bamboogle & HotpotQA & Musique & PopQA & Avg. EM & Avg. Turns \\
\midrule
\multirow{4}{*}{Search-R1}  
 & fixed & fixed & \textbf{70.55} & 47.00 & \textbf{57.92} & \textbf{27.02} & \textbf{42.80} & \textbf{49.49} & 1.72\\
 & fixed & 18 & 45.95 & \textbf{49.00} & 50.77 & 21.47 & 40.25 & 39.89 & 1.74\\
 & 18 & fixed & 60.77 & 45.20 & 51.12 & 20.07 & 42.60 & 43.69 & 2.23\\
 & 18 & 18    & 43.82 & 46.60 & 44.62 & 15.77 & 39.10 & 36.15 & 2.31 \\
\midrule
\multirow{4}{*}{GiGPO}  
 & fixed & fixed & \textbf{67.95} & \textbf{52.00} & \textbf{53.37} & \textbf{24.37} & \textbf{42.65} & \textbf{47.23} & 1.62 \\
 & fixed & 18 & 43.00 & 50.20 & 46.27 & 20.40 & 39.82 & 37.76 & 1.67 \\
 & 18 & fixed & 63.37 & 47.80 & 50.02 & 20.02 & 42.57 & 44.11 & 1.38 \\
 & 18 & 18    & 44.72 & 49.40 & 45.12 & 16.25 & 39.77 & 36.86 & 1.38 \\
\midrule
\multirow{4}{*}{IGPO}  
 & fixed & fixed & \textbf{70.32} & \textbf{47.40} & \textbf{56.72} & \textbf{25.05} & \textbf{44.62} & \textbf{49.12} & 2.99\\
 & fixed & 18 & 49.45 & 46.80 & 48.90 & 21.27 & 41.77 & 40.54 & 3.00 \\
 & 18 & fixed & 62.12 & 45.60 & 50.62 & 19.82 & 42.00 & 43.70 & 1.67 \\
 & 18 & 18    & 43.62 & 46.60 & 44.62 & 16.65 & 39.60 & 36.44 & 1.71 \\
\midrule
\multirow{4}{*}{Tree-GRPO}  
 & fixed & fixed & \textbf{65.27} & 43.80 & \textbf{51.15} & \textbf{21.87} & 40.32 & \textbf{44.63} & 2.07 \\
 & fixed & 18 & 45.22 & \textbf{45.40} & 42.12 & 16.70 & 37.15 & 35.60 & 2.08 \\
 & 18 & fixed & 58.35 & 39.80 & 49.15 & 20.20 & \textbf{42.95} & 42.57 & 1.79 \\
 & 18 & 18    & 43.42 & 39.60 & 42.20 & 16.87 & 40.07 & 35.76 & 1.77 \\
\bottomrule
\end{tabular}}
\caption{Performance comparison across five datasets on different train/test environments. The \textbf{best results} within a comparing group are bold. Avg. Turns means the average search turns called by the agent. Fixed and 18 refer to the Wiki-fixed and Wiki-18, respectively.}
\label{tab:evn_detailed}
\end{table*}

Corresponding to Tab.~\ref{tab:compare_corpus}, we report the detailed results in Tab.~\ref{tab:corpus_detailed}.

\begin{table*}[htbp]
\centering
\resizebox{\textwidth}{!}{\begin{tabular}{ccccccccc}
\toprule
Method& Train Env. & 2Wiki & Bamboogle & HotpotQA & Musique & PopQA & Avg. EM & Avg. Turns \\
\midrule
\multirow{2}{*}{Search-R1}  
 & fixed 
 & 63.30 & \textbf{51.00} & 52.90 & \textbf{22.95}& \textbf{43.45} & 45.81 & 2.11\\
 & 18 
 & \textbf{64.90} & 44.40 & \textbf{54.37} & 21.95 & 43.37 & \textbf{46.09} & 1.72\\
\midrule
\multirow{2}{*}{GiGPO}  
 & fixed 
 & 66.20 & \textbf{51.20} & \textbf{52.30} & \textbf{22.37} & 42.60 & \textbf{46.03} & 1.64 \\
 & 18 
 & \textbf{67.47} & 51.00 & 50.92 & 21.45 & \textbf{42.77} & 45.81 & 1.46 \\
\midrule
\multirow{2}{*}{IGPO}  
 & fixed 
 & \textbf{67.52} & 49.60 & 54.20 & 22.07 & \textbf{43.22} & 46.84 & 2.20\\
 & 18
 & 67.15 & \textbf{52.60} & \textbf{56.27} & \textbf{25.35} & 42.02 & \textbf{47.87} & 1.63 \\
\midrule
\multirow{2}{*}{Tree-GRPO}  
 & fixed 
 & 63.47 & 42.60 & 49.50 & 20.15 & 43.07 & 44.00 & 1.86 \\
 & 18 
 & \textbf{65.42} & \textbf{45.80} & \textbf{50.47} & \textbf{20.40} & 43.15 & \textbf{44.89} & 1.75 \\
\bottomrule
\end{tabular}}
\caption{Performance comparison across five datasets on different train environments. All the methods are tested on the Wiki-Fixed environment. The \textbf{best results} within a comparing group are bold. Avg. Turns means the average search turns called by the agent. Fixed and 18 refer to the Wiki-fixed and Wiki-18, respectively.}
\label{tab:corpus_detailed}
\end{table*}

Corresponding to Fig.~\ref{fig:curves}, we report the detailed results in Tab.~\ref{tab:data_detailed}.

\begin{table*}[htbp]
\centering
\resizebox{\textwidth}{!}{\begin{tabular}{ccccccccc}
\toprule
Method& Train Data & 2Wiki & Bamboogle & HotpotQA & Musique & PopQA & Avg. EM & Avg. Turns \\
\midrule
\multirow{3}{*}{Search-R1}  
 & 9,000
 & 63.50 & \textbf{44.80} & \textbf{52.57} & \textbf{25.40} & \textbf{43.82} & \textbf{46.27} & 2.97 \\
 & 15,000 
 & \textbf{64.15} & 43.60 & 51.97 & 23.05 & 41.90 & 45.21 & 3.00 \\
 & 45,000
 & 60.10 & 35.60 & 50.72 & 23.02 & 40.50 & 43.34 & 2.09 \\
\midrule
\multirow{3}{*}{GiGPO}  
 & 9,000
 & \textbf{58.47} & \textbf{44.40} & 50.58 & 19.57 & \textbf{44.70} & 43.36 & 1.47\\
 & 15,000 
 & 53.20 & 41.80 & 50.10 & \textbf{23.97} & 41.85 & 42.26 & 1.62\\
 & 45,000
 & 58.05 & 40.80 & \textbf{51.70} & 23.35 & 43.70 & \textbf{44.09} & 1.59\\
\midrule
\multirow{3}{*}{IGPO}  
 & 9,000
 & 32.25 & 9.20 & 15.90 & 5.18 & 5.70 & 14.58 & 3.00\\
 & 15,000 
 & \textbf{50.75} & \textbf{34.80} &\textbf{48.45} & \textbf{18.77} & 41.12 & \textbf{39.62} & 2.16\\
 & 45,000
 & 48.60 & 33.40 & 46.10 & 18.42 & \textbf{42.07} & 38.63 & 2.00\\
\midrule
\multirow{3}{*}{Tree-GRPO}  
 & 9,000
 & 57.20 & 37.60 & 44.80 & 21.90 & 40.57 & 41.01 & 3.00\\
 & 15,000 
 & 52.32 & 34.40 & 41.20 & 18.07 & 31.05 & 35.62 & 2.39\\
 & 45,000
 & \textbf{62.60} & \textbf{38.60} & \textbf{48.82} & \textbf{22.20} & \textbf{41.27} & \textbf{43.57} & 2.96 \\
\bottomrule
\end{tabular}}
\caption{Performance comparison across five datasets on different training data. The \textbf{best results} within a comparing group are bold. Avg. Turns means the average search turns called by the agent.}
\label{tab:data_detailed}
\end{table*}

Corresponding to Fig.~\ref{fig:batchsizes}, we report the detailed results in Tab.~\ref{tab:bsz_3} and Tab.~\ref{tab:bsz_2} for Qwen3-8B with Hermes format and Qwen2.5-7B-Instruct with Search-R1 format, respectively.

\begin{table*}[htbp]
\centering
\resizebox{\textwidth}{!}{\begin{tabular}{ccccccccc}
\toprule
Method& batchsize & 2Wiki & Bamboogle & HotpotQA & Musique & PopQA & Avg. EM & Avg. Turns \\
\midrule
\multirow{5}{*}{Search-R1}  
 & 32
 & \textbf{70.55} & 47.00 & \textbf{57.92} & \textbf{27.02} & 42.80 & \textbf{49.49} & 1.72\\
 & 64 
 & 69.25 & 47.60 & 55.05 & 23.07 & 43.22 & 47.64
 & 2.05 \\
 & 128
 & 68.57 & \textbf{50.60} & 55.65 & 26.72 & 43.77 & 48.73
 & 1.85 \\
 & 256
 & 63.92 & 46.80 & 50.65 & 21.42 & \textbf{43.92} & 45.03
 & 1.63 \\
 & 512
 & 63.08 & 40.20 & 48.81 & 18.27 & 42.47 & 43.07
 & 1.79 \\
\midrule
\multirow{5}{*}{GiGPO}  
 & 32
 & \textbf{67.95} & \textbf{52.00} & \textbf{53.37} & \textbf{24.37} & 42.65 & \textbf{47.23} & 1.62  \\
 & 64 
 & 66.17 & 49.20 & 52.87 & 20.60 & \textbf{44.25} & 46.07
 & 1.44 \\
 & 128
 & 63.42 & 44.00 & 48.32 & 18.00 & 40.60 & 42.63
 & 1.51 \\
 & 256
 & 59.67 & 44.60 & 47.67 & 18.45 & 41.17 & 41.83
 & 1.31 \\
 & 512
 & 58.05 & 40.00 & 44.77 & 14.00 & 37.95 & 38.73
 & 1.38 \\
\midrule
\multirow{5}{*}{IGPO}  
 & 32
 & \textbf{70.32} & 47.40 & \textbf{56.72} & \textbf{25.05} & \textbf{44.62} & \textbf{49.12} & 2.99 \\
 & 64 
 & 65.35 & \textbf{48.20} & 49.90 & 18.00 & 42.70 & 44.11
 & 1.47 \\
 & 128
 & 64.12 & 46.20 & 50.02 & 20.20 & 41.52 & 44.03
 & 1.56 \\
 & 256
 & 63.42 & 40.40 & 49.17 & 19.57 & 41.07 & 43.22
 & 1.54 \\
 & 512
 & 59.97 & 43.60 & 44.92 & 14.87 & 38.52 & 39.69
 & 1.53 \\
\midrule
\multirow{5}{*}{Tree-GRPO}  
 & 32
 & 65.27 & 43.80 & \textbf{51.15} & \textbf{21.87} & 40.32 & \textbf{44.63} & 2.07 \\
 & 64 
 & 63.02 & 43.00 & 50.17 & 19.15 & \textbf{42.35} & 43.65
 & 2.06 \\
 & 128
 & \textbf{66.67} & \textbf{44.60} & 49.12 & 19.92 & 41.80 & 44.38
 & 1.95 \\
 & 256
 & 62.95 & 42.40 & 49.97 & 18.62 & 41.82 & 43.31
 & 1.81 \\
 & 512
 & 63.27 & 42.80 & 47.87 & 18.40 & 41.42 & 42.74
 & 1.59 \\
\bottomrule
\end{tabular}}
\caption{Performance comparison across five datasets on different training batch sizes, with the Qwen3-8B as the base model and Hermes format tool call. The \textbf{best results} within a comparing group are bold. Avg. Turns means the average search turns called by the agent.}
\label{tab:bsz_3}
\end{table*}

\begin{table*}[htbp]
\centering
\resizebox{\textwidth}{!}{\begin{tabular}{ccccccccc}
\toprule
Method& batchsize & 2Wiki & Bamboogle & HotpotQA & Musique & PopQA & Avg. EM & Avg. Turns \\
\midrule
\multirow{5}{*}{Search-R1}  
 & 32
 & 57.32 & 40.80 & 47.67 & 18.22 & 35.02 & 39.60 
 & 2.11 \\
 & 64 
 & 62.25 & 41.80 & 52.15 & 23.55 & 42.17 & 44.93
 & 2.98 \\
 & 128
 & 63.77 & 43.00 & 51.8 & 23.28 & 40.86 & 44.87
 & 2.02 \\
 & 256
 & 51.4 & 40.60 & 48.87 & 20.25 & 41.47 & 40.50
 & 2.88 \\
 & 512
 & 57.47 & 36.00 & 47.73 & 15.62 & 37.62 & 39.50
 & 2.99 \\
\midrule
\multirow{5}{*}{GiGPO}  
 & 32
 & 53.17 & 38.80 & 48.17 & \textbf{21.50} & \textbf{42.30} & 41.21 
 & 1.67 \\
 & 64 
 & 61.25 & 40.60 & 49.52 & 21.45 & 41.15 & 43.26
 & 1.62 \\
 & 128
 & 51.47 & 43.40 & 48.32 & 21.47 & 39.65 & 40.32
 & 1.75 \\
 & 256
 & 53.72 & 42.20 & 49.30 & 21.15 & 42.32 & 41.64
 & 1.78 \\
 & 512
 & 57.5 & 40.40 & 49.00 & 16.3 & 35.72 & 39.65
 & 2.14 \\
\midrule
\multirow{5}{*}{IGPO}  
 & 32
 & 47.35 & 39.60 & 48.25 & 18.32 & 38.45 & 38.13 
 & 2.01 \\
 & 64 
 & 58.1 & 42.2 & 49.72 & 19.97 & 42.5 & 42.56
 & 1.70 \\
 & 128
 & 51.25 & 38 & 48.075 & 18.475 & 41.2 & 39.6969
 & 2.82 \\
 & 256
 & 48.32 & 32.6 & 47.07 & 13.57 & 42.2 & 37.63
 & 2.13 \\
 & 512
 & 52.72 & 36.2 & 44.32 & 15.1 & 37.75 & 37.43
 & 2.00 \\
\midrule
\multirow{5}{*}{Tree-GRPO}  
 & 32
 & 59.75 & 39.00 & 48.60 & 17.70 & 39.30 & 41.26 
 & 1.88 \\
 & 64 
 & 55.32 & 32.60 & 45.27 & 17.57 & 35.55 & 38.25
 & 2.20 \\
 & 128
 & 55.66 & 40.25 & 48.46 & 19.60 & 38.49 & 40.54
 & 2.57 \\
 & 256
 & 55.95 & 40.40 & 48.72 & 16.90 & 38.37 & 40.00
 & 2.87 \\
 & 512
 & 55.77 & 41.00 & 48.45 & 17.20 & 37.05 & 39.66
 & 2.29 \\
\bottomrule
\end{tabular}}
\caption{Performance comparison across five datasets on different training batch sizes, with the Qwen3-8B as the base model and Hermes format tool call. The \textbf{best results} within a comparing group are bold. Avg. Turns means the average search turns called by the agent.}
\label{tab:bsz_2}
\end{table*}

Corresponding to Fig.~\ref{fig:max_turns}, we report the detailed results in Tab.~\ref{tab:searchr1_turn}, Tab~\ref{tab:gigpo_turn}, Tab.~\ref{tab:igpo_turn}, and Tab.~\ref{tab:tree_turn} for Search-R1, GiGPO, IGPO, and Tree-GRPO, respectively.

\begin{table*}[htbp]
\centering
\resizebox{\textwidth}{!}{\begin{tabular}{ccccccccc}
\toprule
Train Budget& Test Budget & 2Wiki & Bamboogle & HotpotQA & Musique & PopQA & Avg. EM & Avg. Turns \\
\midrule
\multirow{4}{*}{4}  
 & 4
 & \textbf{70.55} & 47.00 & 57.92 & \textbf{27.02} & 42.80 & 49.49 
 & 1.72\\
 & 8 
 & 70.45 & \textbf{47.80} & 57.90 & \textbf{28.17} & 42.27 & \textbf{49.64}
 & 1.83 \\
 & 16
 & 69.87 & 47.00 & 57.07 & 27.97 & 42.60 & 49.30
 & 1.73 \\
 & 32
 & 70.3 & 48.60 & 57.32 & 27.92 & 42.55 & 49.49
 & 1.73 \\
\midrule
\multirow{4}{*}{8}  
 & 4
 & 33.80 & 25.60 & 25.35 & 9.15 & 15.02 & 20.97
 & 2.96\\
 & 8 
 & 65.32 & 45.60 & 50.85 & \textbf{21.25} & 39.60 & 44.29
 & 4.00 \\
 & 16
 & \textbf{65.87} & \textbf{47.80} & 51.15 & \textbf{21.25} & 40.47 & 44.78
 & 4.02 \\
 & 32
 & 65.72 & 46.40 & \textbf{51.60} & 21.15 & \textbf{40.55} & \textbf{44.80}
 & 4.01 \\
\midrule
\multirow{4}{*}{16}  
 & 4
 & 48.10 & 39.00 & \textbf{47.10} & 16.00 & 32.27 & 35.96
 & 2.35\\
 & 8 
 & 56.77 & 37.20 & 46.95 & 15.97 & \textbf{32.32} & \textbf{37.98}
 & 2.52 \\
 & 16
 & 56.32 & 39.00 & 46.62 & \textbf{16.27} & 32.15 & 37.87
 & 2.51 \\
 & 32
 & \textbf{57.10} & \textbf{40.60} & 46.27 & 16.15 & 31.62 & 37.87
 & 2.52 \\
\midrule
\multirow{4}{*}{32}  
 & 4
 & 27.97 & 23.00 & 29.05 & 9.50 & 21.87 & 22.12
 & 2.74\\
 & 8 
 & 56.27 & 42.00 & 45.10 & 18.72 & 36.00 & 39.11
 & 4.72 \\
 & 16
 & 60.20 & 44.00 & 49.22 & 22.52 & 39.27 & 42.84
 & 5.40 \\
 & 32
 & \textbf{61.05} & \textbf{46.00} & \textbf{49.35} & \textbf{22.95} & \textbf{39.33} & \textbf{43.25}
 & 5.40 \\
\bottomrule
\end{tabular}}
\caption{\textbf{Search-R1} performance comparison across five datasets on different training/test search budgets, with the Qwen3-8B as the base model and Hermes format tool call. The \textbf{best results} within a comparing group are bold. Avg. Turns means the average search turns called by the agent.}
\label{tab:searchr1_turn}
\end{table*}

\begin{table*}[htbp]
\centering
\resizebox{\textwidth}{!}{\begin{tabular}{ccccccccc}
\toprule
Train Budget& Test Budget & 2Wiki & Bamboogle & HotpotQA & Musique & PopQA & Avg. EM & Avg. Turns \\
\midrule
\multirow{4}{*}{4}  
 & 4
 & 67.95 & 52.00 & 53.37 & 24.37 & 42.65 & 47.23
 & 1.62\\
 & 8 
 & 67.87 & 50.20 & 53.32 & 25.17 & 42.87 & 47.40
 & 1.62 \\
 & 16
 & \textbf{68.25} & 52.00 & \textbf{53.55} & 25.20 & 42.42 & \textbf{47.49}
 & 1.61 \\
 & 32
 & 67.82 & \textbf{52.80} & 52.70 & \textbf{25.25} & \textbf{42.60} & 47.26
 & 1.61 \\
\midrule
\multirow{4}{*}{8}  
 & 4
 & 61.45 & 46.60 & \textbf{49.60} & \textbf{18.20} & 43.75 & 43.35
 & 1.42\\
 & 8 
 & \textbf{62.22} & \textbf{48.20} & 49.35 & 18.05 & 43.35 & \textbf{43.39}
 & 1.42 \\
 & 16
 & 61.95 & 46.40 & 48.95 & 18.15 & 44.07 & 43.37
 & 1.42 \\
 & 32
 & 61.77 & 47.20 & 48.47 & \textbf{18.20} & \textbf{44.35} & 43.32
 & 1.42 \\
\midrule
\multirow{4}{*}{16}  
 & 4
 & 65.82 & \textbf{50.20} & 49.35 & 21.20 & 43.05 & 45.01
 & 1.59\\
 & 8 
 & 65.62 & \textbf{50.20} & \textbf{49.62} & \textbf{21.27} & 43.10 & \textbf{45.06}
 & 1.55 \\
 & 16
 & \textbf{66.05} & 49.80 & 49.37 & 20.85 & 43.00 & 44.96
 & 1.55 \\
 & 32
 & 65.42 & 49.60 & 49.32 & 21.17 & \textbf{43.25} & 44.93
 & 1.55 \\
\midrule
\multirow{4}{*}{32}  
 & 4
 & \textbf{61.05} & 46.80 & 49.02 & \textbf{19.42} & \textbf{41.77} & \textbf{42.93}
 & 1.44\\
 & 8 
 & 58.62 & 46.20 & 48.22 & 18.50 & 41.20 & 41.77
 & 1.53 \\
 & 16
 & 60.15 & 43.80 & \textbf{49.15} & 19.02 & 41.17 & 42.41
 & 1.41 \\
 & 32
 & 60.45 & \textbf{47.20} & 47.95 & 19.02 & 41.15 & 42.29
 & 1.41 \\
\bottomrule
\end{tabular}}
\caption{\textbf{GiGPO} performance comparison across five datasets on different training/test search budgets, with the Qwen3-8B as the base model and Hermes format tool call. The \textbf{best results} within a comparing group are bold. Avg. Turns means the average search turns called by the agent.}
\label{tab:gigpo_turn}
\end{table*}

\begin{table*}[htbp]
\centering
\resizebox{\textwidth}{!}{\begin{tabular}{ccccccccc}
\toprule
Train Budget& Test Budget & 2Wiki & Bamboogle & HotpotQA & Musique & PopQA & Avg. EM & Avg. Turns \\
\midrule
\multirow{4}{*}{4}  
 & 4
 & \textbf{70.32} & 47.40 & 56.72 & 25.05 & \textbf{44.62} & \textbf{49.12}
 & 3.00\\
 & 8 
 & 69.52 & \textbf{50.00} & 56.32 & 24.77 & 44.10 & 48.72
 & 3.00 \\
 & 16
 & 69.37 & 47.20 & \textbf{56.90} & 24.90 & 44.25 & 48.80
 & 3.00 \\
 & 32
 & 69.40 & 47.60 & 56.57 & \textbf{25.20} & 44.57 & 48.89
 & 3.00 \\
\midrule
\multirow{4}{*}{8}  
 & 4
 & 60.10 & 41.40 & 47.95 & 17.10 & 40.55 & 41.42
 & 2.49\\
 & 8 
 & \textbf{66.60} & 45.80 & 51.80 & 19.80 & 42.47 & 45.18
 & 2.70 \\
 & 16
 & 66.57 & \textbf{49.40} & 51.40 & \textbf{20.25} & 43.00 & \textbf{45.43}
 & 2.70 \\
 & 32
 & 66.07 & 47.40 & \textbf{52.17} & 19.35 & \textbf{42.85} & 45.18
 & 2.70 \\
\midrule
\multirow{4}{*}{16}  
 & 4
 & 50.22 & 40.20 & 45.42 & 16.32 & 34.77 & 36.79
 & 2.68\\
 & 8 
 & 60.40 & 43.80 & \textbf{50.10} & 19.75 & 38.32 & 42.19
 & 2.97 \\
 & 16
 & 60.47 & \textbf{46.40} & 49.75 & 19.50 & 38.60 & 42.21
 & 2.96 \\
 & 32
 & \textbf{60.70} & 42.60 & 49.85 & \textbf{20.30} & \textbf{38.62} & \textbf{42.37}
 & 2.95 \\
\midrule
\multirow{4}{*}{32}  
 & 4
 & 48.27 & 46.80 & 50.12 & 18.85 & 40.40 & 39.63
 & 2.03\\
 & 8 
 & 66.70 & 47.40 & \textbf{51.65} & \textbf{20.95} & \textbf{40.57} & \textbf{45.04}
 & 2.16 \\
 & 16
 & 66.20 & 46.40 & 51.27 & 20.22 & 40.55 & 44.61
 & 2.16 \\
 & 32
 & \textbf{66.92} & \textbf{47.60} & 51.20 & 20.67 & 40.42 & 44.89
 & 2.16 \\
\bottomrule
\end{tabular}}
\caption{\textbf{IGPO} performance comparison across five datasets on different training/test search budgets, with the Qwen3-8B as the base model and Hermes format tool call. The \textbf{best results} within a comparing group are bold. Avg. Turns means the average search turns called by the agent.}
\label{tab:igpo_turn}
\end{table*}

\begin{table*}[htbp]
\centering
\resizebox{\textwidth}{!}{\begin{tabular}{ccccccccc}
\toprule
Train Budget& Test Budget & 2Wiki & Bamboogle & HotpotQA & Musique & PopQA & Avg. EM & Avg. Turns \\
\midrule
\multirow{4}{*}{4}  
 & 4
 & 65.27 & 43.80 & \textbf{51.15} & 21.87 & 40.32 & 44.63
 & 2.07\\
 & 8 
 & 65.12 & \textbf{45.20} & 51.10 & 21.65 & 40.25 & 44.55
 & 2.08 \\
 & 16
 & \textbf{65.35} & 44.80 & 50.97 & 21.70 & 40.27 & 44.58
 & 2.08 \\
 & 32
 & 65.02 & 44.00 & 50.55 & \textbf{22.42} & \textbf{40.72} & \textbf{44.66}
 & 2.07 \\
\midrule
\multirow{4}{*}{8}  
 & 4
 & 63.77 & 45.60 & 49.95 & 20.32 & 38.90 & 43.30
 & 1.84\\
 & 8 
 & 64.25 & \textbf{48.40} & 51.92 & 21.10 & 41.25 & 44.74
 & 2.01 \\
 & 16
 & 64.47 & 46.80 & \textbf{52.05} & 21.47 & 40.47 & 44.68
 & 2.04 \\
 & 32
 & \textbf{65.02} & \textbf{48.40} & 51.50 & \textbf{21.62} & \textbf{41.32} & \textbf{44.97}
 & 2.04 \\
\midrule
\multirow{4}{*}{16}  
 & 4
 & \textbf{67.12} & \textbf{53.20} & \textbf{52.07} & \textbf{19.92} & \textbf{43.15} & \textbf{45.80}
 & 1.90\\
 & 8 
 & 64.22 & 47.40 & 49.00 & 18.17 & 40.70 & 43.15
 & 1.70 \\
 & 16
 & 63.72 & 47.80 & 48.72 & 17.87 & 39.80 & 42.69
 & 1.69 \\
 & 32
 & 63.80 & 46.20 & 48.25 & 17.92 & 40.00 & 42.60
 & 1.73 \\
\midrule
\multirow{4}{*}{32}  
 & 4
 & 63.50 & 45.40 & 48.45 & 18.20 & 38.95 & 42.36
 & 2.33\\
 & 8 
 & 64.47 & 46.40 & 50.15 & 20.32 & 40.97 & 44.05
 & 2.55 \\
 & 16
 & \textbf{65.20} & \textbf{47.20} & \textbf{50.22} & 19.62 & 40.87 & 44.07
 & 2.55 \\
 & 32
 & 64.95 & 44.80 & 50.20 & \textbf{20.42} & \textbf{41.53} & \textbf{44.29}
 & 2.54 \\
\bottomrule
\end{tabular}}
\caption{\textbf{Tree-GRPO} performance comparison across five datasets on different training/test search budgets, with the Qwen3-8B as the base model and Hermes format tool call. The \textbf{best results} within a comparing group are bold. Avg. Turns means the average search turns called by the agent.}
\label{tab:tree_turn}
\end{table*}

\section{The Use of LLMs}

This paper employed LLMs solely for grammatical correction and stylistic refinement, with the purpose of more effectively communicating our results and conclusions.

\end{document}